\documentclass{article}

\usepackage{microtype}
\usepackage{graphicx}
\usepackage{subfigure}
\usepackage{booktabs} 

\usepackage{hyperref}

\usepackage[accepted]{icml2024}

\usepackage{enumitem}
\setlist[itemize]{leftmargin=0.5cm,itemsep=0cm,topsep=0cm}

\usepackage{amsmath,amsfonts,bm}

\def\eqref#1{Equation~(\ref{#1})}

\def\1{\bm{1}}

\DeclareMathAlphabet{\mathsfit}{\encodingdefault}{\sfdefault}{m}{sl}
\SetMathAlphabet{\mathsfit}{bold}{\encodingdefault}{\sfdefault}{bx}{n}

\newcommand{\E}{\mathbb{E}}

\newcommand{\KL}{D_{\mathrm{KL}}}

\DeclareMathOperator{\sign}{sign}

\usepackage{hyperref}
\usepackage{url}
\usepackage{amsthm}
\usepackage{subfigure}
\usepackage{graphicx}

\usepackage{bbm}

\usepackage{enumitem}
\setlist[itemize]{leftmargin=0.5cm}

\newcommand{\beq}{\begin{equation}}
\newcommand{\eeq}{\end{equation}}

\newcommand{\beqa}{\begin{eqnarray}}
\newcommand{\eeqa}{\end{eqnarray}}

\newcommand{\beqan}{\begin{eqnarray*}}
\newcommand{\eeqan}{\end{eqnarray*}}

\renewcommand{\P}{{\cal {P}}}
\renewcommand{\Pr}{\mathbb{P}}

\renewcommand{\E}{\mathbb{E}}

\renewcommand{\1}[1]{\mathbb{I}\{#1\}}

\newcommand{\piref}{\mu}
\renewcommand{\KL}{\mbox{KL}}

\newlength{\minipagewidth}
\newcommand{\eqdef}{\stackrel{\rm def}{=}}

\newtheorem{predefinition}{Definition}

\newtheorem{theorem}{Theorem}

\newtheorem{proposition}{Proposition}
\newtheorem{lemma}{Lemma}

\renewcommand{\hat}{\widehat}
\renewcommand{\phi}{\varphi}

\newcommand{\actionspace}{\mathcal{Y}}

\usepackage{xcolor}

\icmltitlerunning{Nash Learning from Human Feedback}

\begin{document}

\twocolumn[
\icmltitle{Nash Learning from Human Feedback}



\icmlsetsymbol{equal}{*}
\icmlsetsymbol{xoog}{$\dagger$}

\begin{icmlauthorlist}
\icmlauthor{R\'emi Munos\!}{equal,gdm}
\icmlauthor{Michal Valko\!}{equal,gdm}
\icmlauthor{Daniele Calandriello\!}{equal,gdm}
\icmlauthor{Mohammad\,Gheshlaghi\,Azar\!}{equal,gdm,xoog}
\icmlauthor{Mark\,Rowland\!}{equal,gdm}
\icmlauthor{Daniel\,Guo\!}{equal,gdm}
\icmlauthor{Yunhao\,Tang\!}{equal,gdm}
\icmlauthor{Matthieu\,Geist\!}{equal,gdm,xoog}
\icmlauthor{Thomas\,Mesnard\!}{gdm}
\icmlauthor{C\^ome\,Fiegel\!}{ensae}
\icmlauthor{Andrea\,Michi\!}{gdm}
\icmlauthor{Marco\,Selvi\!}{gdm}
\icmlauthor{Sertan\,Girgin\!}{gdm}
\icmlauthor{Nikola\,Momchev\!}{gdm}
\icmlauthor{Olivier\,Bachem\!}{gdm}
\icmlauthor{Daniel\,J.\,Mankowitz\!}{gdm}
\icmlauthor{Doina\,Precup\!}{gdm}
\icmlauthor{Bilal\,Piot}{equal,gdm}
\end{icmlauthorlist}

\icmlaffiliation{gdm}{Google~DeepMind}
\icmlaffiliation{ensae}{ENSAE Paris \textsuperscript{$\dagger$}Now at Cohere}

\icmlcorrespondingauthor{Remi Munos}{remi.munos@inria.fr}
\icmlcorrespondingauthor{Michal Valko}{michal.valko@inria.fr}
\icmlcorrespondingauthor{Daniele Calandriello}{dcalandriello@google.com}
\icmlcorrespondingauthor{Bilal Piot}{piot@google.com}

\icmlkeywords{Large Language Models, Fine-tuning, RLHF, Reinforcement Learning, Nash equilibrium, preference models, alignment with human data}

\vskip 0.3in
]



\printAffiliationsAndNotice{\icmlEqualContribution} 

\begin{abstract}
Reinforcement learning from human feedback (RLHF) has emerged as the main paradigm for aligning large language models (LLMs) with human preferences. 
Traditionally, RLHF involves the initial step of learning a reward model from pairwise human feedback, i.e., expressed as preferences between pairs of text generations. Subsequently, the LLM's policy is fine-tuned to maximize the reward through a reinforcement learning algorithm. 

In this study, we introduce an alternative pipeline for the fine-tuning of LLMs using pairwise human feedback. Our approach entails the initial learning of a \textit{pairwise} preference model, which is conditioned on two inputs (instead of a single input in the case of a reward model) given a prompt, followed by the pursuit of a policy that consistently generates responses preferred over those generated by any competing policy, thus defining the \textit{Nash equilibrium} of this preference model. We term this approach \emph{Nash learning from human feedback} (NLHF).

In the context of a tabular policy representation, we present a novel algorithmic  solution, Nash-MD, founded on the principles of mirror descent. This algorithm produces a sequence of policies, with the last iteration converging to the regularized Nash equilibrium. Additionally, we explore parametric representations of policies and introduce gradient descent algorithms for deep-learning architectures.
We illustrate the effectiveness of our approach by presenting experimental results on a text summarization task.
We believe NLHF offers a compelling avenue for fine-tuning LLMs and enhancing the alignment of LLMs with human preferences.
\end{abstract}

\section{Introduction}
\label{sec:intro}
Large language models (LLMs) \citep{glaese2022improving,anil2023palm,openai2023gpt4,InstructGPT} have made remarkable strides in enhancing natural language understanding and generation. Their success in conversational applications often relies on aligning these models with human preferences, a process primarily guided by the paradigm of reinforcement learning from human feedback (RLHF). A prevailing approach within RLHF involves the initial step of constructing a reward model based on pairwise human preferences, frequently employing the Bradley-Terry model \citep[BT; ][]{Bradley-Terry}.
This reward model assigns an individual score to each generation of the language model conditioned on a given prompt, akin to how the Elo ranking system \citep{elo} assigns scores to chess players to estimate their relative strengths.
Subsequently, model refinement takes place by optimizing the LLM's performance with respect to this reward model through reinforcement learning (RL) over sampled text generations.

However, this BT model has its limitations, primarily coming from its inability to accommodate the full spectrum of possible preferences.
For example, \citet{Bertrand2023} show the limitations of the Elo model by illustrating where Elo score alone cannot predict the right preferences, even in transitive situations. 
There are also situations where maximizing the Elo score is not aligned with maximizing the probability of winning against the corresponding population of players, even when the preference model can be perfectly expressed using a BT model (see Appendix~\ref{sec:Elo.vs.winning.proba} for an example). 
These observations highlight the necessity for a more profound understanding of the implications of BT-based reward maximization in RLHF for achieving genuine alignment with human preferences.

\paragraph{The NLHF approach:}
In this paper, we introduce an alternative pipeline for fine-tuning LLMs from human preference data, which we term {\bf Nash learning from human feedback} (NLHF). In this framework, we depart from the conventional approach of learning a reward model and instead focus on learning a {\bf preference model} and define our objective to compute the {\bf Nash equilibrium} of this preference model.

The preference model takes two responses, denoted as $y$ and $y'$ (possibly conditioned on a prompt $x$), as input and produces a preference score $\P(y\succ y'|x)$, indicating the preference of response $y$ over response $y'$ given the context $x$. We may think of $\P(y\succ y'|x)$ as the probability that a randomly chosen human prefers response $y$ to response $y'$ given $x$. 

In order to learn a preference model, we can initialize it using AI-feedback by leveraging a LLM prompted in a manner akin to how humans have been asked for their preference, such as by instructing the LLM to generate a 1-vs-2 comparison in response to a prompt like: ``Given $x$, which answer do you prefer, answer 1: $y$ or answer 2: $y'$?''. This initial preference model can be further refined, through supervised learning, by aligning it with human preference data.

Notably, such a learnt preference model does not make any Bradley-Terry assumption thus has the potential to capture the diversity and richness of human preferences contained in the training data. 

Moreover, in contrast to the traditional RLHF setting where the reward model depends on the distribution of data that has been used to train the model, a preference model remains essentially invariant to this data distribution. The main reason why preference models are less sensitive to the data distribution than reward models is that preference models takes as input the two responses to be compared whereas the reward model makes an implicit comparison between the (single) input to the distribution it has been trained on (see Section~\ref{sec:sensitivity.distribution}). 

Once the preference model is established, our primary objective is to calculate the corresponding Nash equilibrium. This equilibrium represents a policy that consistently produces responses preferred, as determined by the preference model, over responses generated by any alternative policy. 

The three key properties of our approach, namely, the ability of the preference model to capture a wide spectrum of human preferences (contained in the data), its lower sensitivity on the data distribution, and the potential for the Nash equilibrium to provide a better alignment with the diversity of human preferences, mark a substantial departure from the conventional RLHF framework. We discuss these properties in greater detail in Section~\ref{sec:r.vs.P.model}.

\paragraph{Practical algorithms:}
To approximate the Nash equilibrium of the two-player game in which actions are responses, and payoffs are specified by the preference model, we employ a deep reinforcement learning algorithm. Given a prompt $x$, we generate two responses, denoted as $y$ and $y'$. The first response, $y$, is generated under the current policy $\pi_\theta$ that we are in the process of optimizing. In contrast, the second response, $y'$, is produced by an alternative policy $\pi'$, which we implement in two different versions: {\bf Nash-MD} and {\bf Nash-EMA} (further elaboration on these versions will be provided below). Nash-MD defines the alternative policy $\pi'$ as a geometric mixture between the initial and the current policy (motivated by mirror descent), whereas Nash-EMA implements a first-order approximation of an exponential moving average (EMA) mixture of past policies. Then, the preference model computes $\P(y\succ y'|x)$, and this preference signal serves as a reward for optimizing our policy $\pi_\theta$ using a (regularized) policy gradient algorithm, as outlined in~\cite{geist2019theory}.

\paragraph{Our contributions:}
Our contributions in this work can be summarized as follows. First, we introduce the concept of Nash learning from human feedback (NLHF), framing it as the task of computing the Nash equilibrium for a general preference model. We proceed by introducing and defining a regularized variant of the preference model. We also establish the existence and uniqueness of the corresponding Nash equilibrium in this context. Then, we consider the case of tabular policy representations and introduce a novel algorithm named Nash-MD. This algorithm, founded on the principles of mirror descent (MD) possesses two important properties. First, it converges to the Nash equilibrium, with the final iteration reaching this equilibrium. This differs from conventional regret-minimization-based algorithms, where it is typically the mixture of past policies that converges, necessitating the storage of past policies. Secondly, Nash-MD learns by competing against alternative policies $\pi'$ that represent a (geometric) mixture between the current policy $\pi_\theta$ and the initial policy. Importantly, this can be accomplished without the need to retain intermediate policies, a feature of particular significance in the context of LLMs with their substantial memory requirements. Additionally, we introduce Nash-EMA, a variation inspired by fictitious play, which uses an exponential moving average of past policy parameters. We introduce policy-gradient algorithms for deep learning architectures, Nash-MD-PG and Nash-EMA-PG, inspired by the tabular algorithms.  We present the results of numerical experiments conducted on a text summarizing task utilizing the TL;DR dataset \citep{TLDR}. In these experiments, we employ the NLHF approach to train several models. To assess their performance, we conduct a pairwise evaluation (using the PaLM 2 Large LLM) of the performance of the models and include a comparison to an RLHF baseline.
We conclude that NLHF opens up new promising directions for aligning LLMs with human preferences. 

\section{Prior work}

\paragraph{Preference-based RL.}
Our contribution falls into a broader area of preference-based RL, where we directly learn from pairwise human preferences instead of a hand-designed or learned scalar reward (see, e.g., the survey by~\citealp{JMLR:v18:16-634}). The canonical form of RLHF was proposed in \cite{DeepRL_from_human_pref2023} and popularized by \cite{chatgpt}, in which one learns a scalar reward model from the preference feedback, followed by policy optimization against the reward model. However, an advantage of directly optimizing for preferences rather than a learnt scalar reward function is the potential to avoid \emph{reward hacking}~\citep{amodei2016concrete}, when agents find a way to maximize a reward without performing what was intended. Furthermore, in domains such as medical applications, it may not only be challenging but also undesirable to provide a single scalar reward.

In general, the preference feedback can be provided in different ways, e.g., on the level of states, actions, or a full trajectory. In this work, we focus on the \textit{trajectory feedback} where the experts provide feedback by selecting the preferred one of the two proposed trajectories. Such a simple form of pairwise feedback is the easiest to implement, and has seen applications in summarization ~\citep{stiennon2020learning}, question-answering~\citep{nakano2021webgpt,menick2022teaching} and general language-based assistants~\citep{InstructGPT,glaese2022improving,bai2022training}. 
Ranking based algorithms in the RLHF literature include the RAFT \cite{raft} and ReST \cite{rest} approaches.
More complicated forms of feedback has been studied in theoretical literature such the work of \citet{efroni2021reinforcement}.

\paragraph{Theoretical guarantees for learning from preferences.}
Learning policies from  preference feedback of histories was studied by~\citet{10.1007/978-3-642-23780-5_11} who learned the preference model for histories and by~\citet{10.1007/978-3-642-23780-5_30} who trained a model ranking
actions for a state. \citet{robi13,BusaFeketeSzorenyiWengChengHullermeier13} approached this setting by comparing and ranking policies and \citet{NIPS2012_16c222aa} by learning a distribution over policy space. 
Preference-based RL is also explored in dueling RL~\citep{pmlr-v124-novoseller20a,pacchiano2023dueling}, which generalizes the well-studied \textit{dueling bandits} problem. In particular, \citet{pacchiano2023dueling} assumes a Bradley-Terry model, which they estimate using maximum likelihood in the tabular setting. 

Our work is also related to results of~\citet{wang2023rlhf}
who consider learning Nash equilibria of the human preference model, and reduce the problem to finding Nash equilibria for a special class of factored two-player Markov games under a restricted set of policies.
The interaction of Nash equilibria and LLMs have been also 
explored by \citet{jacob2023regularized,jacob2023consensus}. 
Moreover, \citet{chen2022humanintheloop} gave 
first results for function approximation in preference-based RL, however with a computationally inefficient algorithm.

\paragraph{Optimization without reward function.} A number of recent works has attempted to optimize for preference feedback without learning a reward function. For example, Direct Preference Optimization \citep[DPO; ][]{DPO} optimizes the policy through a loss function defined via the Bradley-Terry reward model. SLiC~\citep{zhao2023slichf} modifies the classical RLHF training loss by calibrating a ranking loss which contrasts a positive and a negative sequence. This resembles directly optimizing for the pairwise preference, albeit without convergence guarantees. Identity Preference Optimization \citep[IPO; ][]{azar2023general} and the Generalized Preference Optimization \citep[GPO; ][]{tang2024generalized} proposed to directly optimize the pairwise human preference with offline preference data by optimizing against a fixed opponent. And recently, it has been observed that the online version of IPO \citep{calandriello2024human} approximates the Nash equilibrium of a preference model using a particular case of Nash-MD (called Self-Play).

\section{The preference model and its Nash equilibrium}

We now introduce the core conceptual ideas behind our approach to learning from preference feedback. We consider a preference model in a contextual bandit setting. Given a context (or prompt) $x$ in the context space $\mathcal{X}$ and two actions (or responses/choices) $y$ and $y'$ in the action space $\actionspace$, the preference of $y$ over $y'$ given $x$ is a number between $0$ and $1$ which is written ${\cal P}(y\succ y'|x)$. We will assume that the preference model is antisymmetric: $\P(y\succ y'|x) = 1- \P(y'\succ y|x)$.

\newcommand{\randomoutcome}{Z}
\newcommand{\randomoutcomedist}{\nu}



In the context of LLMs we can think of the preference $\P(y\succ y'|x)$ as the probability that a randomly chosen human prefers a response $y$ over the other response $y'$ given the context $x$. 

We define the preference between two distributions conditioned on a state $x$: 
$${\cal P}(\pi\succ\pi'|x)\eqdef \E_{y\sim\pi(\cdot|x), y'\sim\pi'(\cdot|x)}\left[\P(y\succ y'|x)\right]$$ 
and the preference of an action over a distribution ${\cal P}(y \succ\pi'|x)\eqdef \E_{y'\sim\pi'(\cdot|x)}\left[\P(y\succ y'|x)\right]$. Finally, given a distribution $\rho$ over contexts, we define the preference between two policies:
$${\cal P}(\pi\succ\pi')\eqdef\E_{x\sim\rho}\E_{y\sim\pi(\cdot|x), y'\sim\pi'(\cdot|x)}\left[\P(y\succ y'|x)\right].$$
We say that a policy $\pi$ is preferred over (or simply wins against) another policy $\pi'$ if ${\cal P}(\pi\succ\pi') \geq 1/2$.
In the remainder of the paper, we assume (without loss of generality) that $\rho$ assigns every context positive probability.

In this paper we will consider the objective of finding a policy $\pi^*$ which is preferred over any other alternative policy:
\begin{align}\label{eq:minimax}
    \pi^*\eqdef \arg\max_{\pi}\min_{\pi'} {\cal P}(\pi\succ\pi') \, .
\end{align}

This objective implicitly defines a two-player game, in which the players select policies $\pi$ and $\pi'$, the first player receiving a payoff of ${\cal P}(\pi\succ\pi')$, and the second player receiving ${\cal P}(\pi'\succ\pi) = 1 - {\cal P}(\pi\succ\pi')$. This is therefore a two-player, antisymmetric, constant-sum game, and it follows that when both players use a policy $\pi^*$ solving \eqref{eq:minimax}, this is a \emph{Nash equilibrium} for this game, by the minimax theorem \citep{vonNeumann:1928:TGG}. This is the fundamental solution concept we study in this paper.

The objective introduced in \eqref{eq:minimax} has two central differences relative to the majority of existing work on RLHF. First, the objective is expressed directly in terms of preferences themselves, not in terms of a reward function learnt from preferences, and also not in terms of a non-linear transformation of the preferences. Second, our solution concept relies on the notion of Nash equilibrium, rather than on optimization against a fixed behavior. We discuss the impact of both of these choices through several examples below.

\label{sec:r.vs.P.model}

\subsection{Limited expressivity of reward models}

A learnt preference model possesses the capacity to encompass any property of human preferences as much as they are contained in the dataset used to train the model. For example they can model non-transitive preferences (see the examples in Appendix~\ref{apx:non-transitive}), a characteristic not attainable by reward models since they inherently assign a single score to each policy. Whether humans exhibit non-transitive preferences or not has been a subject of longstanding research (see, for instance, \citep{Tversky,Klimenko}). But even if single individuals are transitive, it is nevertheless possible that the resulting expected preference model is not (see Appendix~\ref{apx:non-transitive.2} for an example).

Additionally, non-transitivity is not the only limitation of Bradley-Terry-based reward models; see, e.g., Example~3 in \citep{Bertrand2023} where Elo score fails to capture the correct preference ordering between policies, even in transitive situations. In fact, we show in Appendix~\ref{sec:Elo.vs.winning.proba} that even when the preference model is perfectly captured by the Bradley-Terry model, optimization of the reward/Elo score may \emph{still} disagree with any reasonable notion of preference optimization. 

Therefore, we can safely argue that preference models offer a more flexible and nuanced framework for modeling preferences than BT-based preference models models.

\subsection{Alignment with diversity of human preferences}

Here, we illustrate that in some situations, the solution offered by the Nash equilibrium of the preference model (which we refer to as the NLHF solution) is more aligned with the diversity of human preferences than the optimum of the reward model (which we refer to as the RLHF solution).

Consider the following situation where there are 3 different actions ($y_1$, $y_2$, $y_3$) and we have a population composed of 3 types of humans with respective preferences $\P_1, \P_2, \P_3$, defined in the following way: 
$\P_i(y_1\succ y_2)=\P_i( y_1\succ y_3)=\P_i( y_2\succ y_3)=1/2$, for $1\leq i\leq 3$, except for the following cases: $\P_1( y_2\succ y_1)=1$ (thus $\P_1( y_1\succ y_2)=0$), $\P_2( y_1\succ y_3)=1$ (thus $\P_2( y_3\succ y_1)=0$), and $\P_3( y_3\succ y_2)=1$ (thus $\P_3( y_2\succ y_3)=0$).

Now, let us assume these 3 types form a near-uniform distribution over humans, for example ${\mathbb P}(\mbox{Type } 1)=1/3-\epsilon$, ${\mathbb P}(\mbox{Type } 2)={\mathbb P}(\mbox{Type } 3)=1/3+\epsilon/2$. The corresponding population preference is thus $\P_\epsilon = (1/3-\epsilon)\P_1+(1/3+\epsilon/2)(\P_2+\P_3)$. In the case $\epsilon > 0$ (so Type 1 is slightly less frequent than the other types) then a reward model will assign a slightly better reward (assuming a Bradley-Terry model) to action $y_1$, thus optimizing the expected reward (the RLHF solution) will produce a deterministic policy choosing exclusively $y_1$.

However, here we are in a situation where the preferences are not uniformly aligned across humans \citep{moskovitz2023confronting}. In the case of uniform sampling of humans (i.e., $\epsilon=0$), the Nash equilibrium of $\P_{\epsilon=0}$ is a uniform mixture between the 3 policies. Actually, the preference model $\P_\epsilon$ corresponding to any $\epsilon$ is defined as: $\P_{\epsilon}( y_2\succ y_1)=2/3-\epsilon/2$, $\P_{\epsilon}( y_3\succ y_1)=1/3-\epsilon/4$, $\P_{\epsilon}( y_3\succ y_2)=2/3+\epsilon/4$, $\P_{\epsilon}( y_i\succ y_i)=1/2$, and  $\P_{\epsilon}( y_i\succ y_j)=1 -\P_{\epsilon}( y_j\succ y_i)$, for $1\leq i<j\leq 3$. By a simple calculation, we deduce that for any $|\epsilon|\leq 1/3$, the Nash equilibrium of this preference model consists in selecting $ y_1$ and $ y_2$ with probability $1/3+\epsilon/2$ each, and $ y_3$ with probability $1/3-\epsilon$.

We believe that in this situation, the Nash solution of the preference model (i.e., the NLHF solution), assigning close to uniform probability to these 3 actions (one being preferred by each category of humans) is more aligned with the diversity of human preferences than the optimum of the reward model (i.e., the RLHF solution), which would deterministically select a single action. Also the Nash equilibrium is less sensitive to the preference distribution, since the corresponding equilibrium is smooth w.r.t.~change in the distribution over types of humans (i.e., when $\epsilon$ varies near $0$), whereas the RLHF solution will switch from selecting exclusively $y_1$ when $\epsilon>0$ to selecting exclusively $y_2$ when $\epsilon<0$.

\subsection{Sensitivity to the data distribution}\label{sec:sensitivity.distribution}
Another difference between reward and preference models is that a reward model depends on the distribution it has been trained on, whereas a preference model essentially does not. 
Indeed, when we learn a reward model we are solving the following optimization problem (in the limit of an infinite amount of data):
$$r^{\pi}\eqdef\arg\max_{r(\cdot,\cdot)}
\E_{
{\tiny
\hspace{-3mm}
\begin{array}{c}
x\sim\rho\\
y,y'\sim \pi(\cdot|x) \\
Z\sim\nu
\end{array}
}}
\hspace{-3mm}
\left[ \log\left(\sigma(r(x,y_w^Z)-r(x,y_l^Z))\right)\right],$$ 
where $y_w^Z$ and $y_l^Z$ are respectively the preferred (and less preferred) response (among $y$ and $y'$) according to a randomly sampled human $Z\sim\nu$, given $x$. 
The (optimal) solution to this problem $r^{\pi}$ depends on the policy $\pi$ that has generated the data. 
Indeed, as mentioned in the introduction (see Section~\ref{sec:intro}), the reward model assigns an Elo score to each individual response, which is defined in terms of a comparison against other responses; thus, it depends on the overall distribution over responses it has been trained on. See Theorem~\ref{thm:data.dependent.reward.models} (Appendix~\ref{sec:data.dependent.reward.models}) for a precise statement.

On the contrary, since the preference model takes two responses as input, the output does not depend directly on the distribution these responses have been sampled from. 
The preference model is simply learnt by supervised learning, where for each $x,y,y'$, the preference model $\P(y\succ y'|x)$ is regressed to the human preference $\1{ y \mbox{ is preferred to }y'\mbox{ given }x}$ using a cross entropy loss:
$$\P^* \eqdef \arg\max_{\P(\cdot\succ\cdot|\cdot)}
\E_{
{\tiny
\hspace{-3mm}
\begin{array}{c}
x\sim\rho\\
y\sim \pi(\cdot|x)\\
y'\sim \pi'(\cdot|x)\\
Z\sim\nu
\end{array}
}}\left[ \log \P(y_w^Z\succ y_l^Z|x)\right],$$
where $y_w^Z$ (resp.~$y_l^Z$) is the preferred (resp.~less preferred) response (among the  responses $y$ and $y'$) given $x$ according to a random human $Z\sim\nu$.
Notice that the optimal solution to this optimization problem is just the  probability that a random human prefers $y$ to $y'$ given $x$: for any $x\in\mbox{supp}(\rho)$, $y\in\mbox{supp}(\pi(\cdot|x))$, $y'\in\mbox{supp}(\pi'(\cdot|x))$,
$$\P^*(y\succ y'|x) = {\mathbb P}_{Z\sim\nu} \left( \mbox{Human } Z \mbox{ prefers }y \mbox{ to } y' \mbox{ given }x\right).$$
This quantity is just a function of $x, y, y'$ and does not depends on how $x$, $y$ and $y'$ have been chosen, thus is independent of $\rho$, $\pi$ or $\pi'$.
Thus in this ideal case of perfect representations, a preference model is insensitive to the data generation distribution ($\rho,\pi, \pi'$) whereas a reward model depends on it.
Now, of course, when using approximate models or a finite amount of data, the learned preference model may still depend on the data distribution as the quality of the approximation  depends on the local quantity of collected data. 

Thus it is our general expectation that the preference model is significantly less reliant on the specific policy that generated the data when compared to the reward model. 

This observation becomes even more important in scenarios where multiple iterations of RLHF/NLHF occur, comprising data collection, constructing a reward/preference model, policy optimization based on the model, and collecting new data following the updated policy. 
In the case of RLHF, the reward model from a prior iteration diverges from the next iteration due to shifts in data distributions, necessitating complete relearning. On the contrary, in the NLHF approach, the preference model can be preserved and further enriched through the introduction of novel data, thereby offering a more seamless and efficient adaptation process.

\section{Regularized preference model}
We now consider a regularized version of the preference model. This is motivated by situations where the preference model is more accurately estimated along responses obtained by following a given \textit{reference} policy. This could be the policy responsible for generating the data used to train the preference model or include situations where it is imperative to ensure that our solution remains close to a known safe policy.
In such cases, we incorporate a penalty mechanism into our preference model, employing \KL-regularization to quantify the divergence between the policy under consideration and a designated reference policy denoted as $\mu$; see  \citep{jaques2019way,stiennon2020learning,InstructGPT} for further details on the role of KL-regularization in RLHF.

The regularized preference between actions $y\sim\pi(\cdot|x), y'\sim\pi'(\cdot|x)$ is defined as 
{\small
$$
{\cal P}_{\tau}^{\pi,\pi'} (y\succ y'|x)\eqdef {\cal P}(y\succ y'|x) - \tau \log\frac{\pi(y|x)}{\mu(y|x)} +\tau \log\frac{\pi'(y'|x)}{\mu(y'|x)},$$
}
and we define accordingly the \KL-regularized preference between policies:
{\small
\begin{eqnarray}
{\cal P}_{\tau} (\pi\succ \pi')
\hspace{-3mm}&\eqdef &\hspace{-3mm}\E_{x\sim\rho, y\sim\pi(\cdot|x), y'\sim\pi'(\cdot|x)}\left[{\cal P}_{\tau}^{\pi,\pi'}(y\succ y'|x)\right] \notag \\
\hspace{-3mm}&=&\hspace{-3mm} \label{eq:minmax.reg}
{\cal P}(\pi\succ \pi') - \tau \KL_\rho(\pi,\mu) +\tau \KL_\rho(\pi', \mu)
\end{eqnarray}
}
where $\KL_\rho(\pi,\mu)\eqdef\E_{x\sim\rho}[ \KL(\pi(\cdot|x),\mu(\cdot|x))]$. 
We now state the existence and uniqueness of the Nash equilibrium of this regularized preference model (see the proof in the Appendix~\ref{proof:unique}):
\begin{proposition}[Nash equilibrium]
\label{prop:unique}
There exists a unique Nash equilibrium of the regularized preference model ${\cal P}_{\tau}$. 
\end{proposition}

\newpage

\section{Algorithms for approximating the Nash eq.}

The regularized preference model $\P_\tau(\pi\succ\pi')$ defines a constant-sum two-player game where Player 1 selects $\pi$ and Player 2 selects $\pi'$. There are well-known techniques for approximating the Nash equilibrium. Some of them offer a convergence on average (in the sense that it is a mixture of the sequence of policies that converges to the Nash equilibrium), whereas other methods offer convergence of the last iterate. 

\paragraph{Convergence on average.}
{\em Fictitious play}  \citep[FP; ][]{brown1951,Robinson1951,pmlr-v37-heinrich15,Fudenberg.Levine1998}
consists in playing, at every iteration, each player's best response against the uniform mixture of the opponent's past strategies. Here we would define $\pi_{t+1}\eqdef \arg\max_\pi \P(\pi\succ\bar\pi_t)$, where $\bar\pi_t$ is the mixture policy $\frac 1t\sum_{s=1}^t \pi_s$. 
It is known that the mixture policy $\bar\pi_t$ converges to the Nash equilibrium in constant-sum games (see \citep{hofbauer2006best} for a reference in the general concave-convex case considered here).
Also, FP has been considered with function approximation~\citep{Heinrich16}.
{\em Online convex optimization: }In the context of solving convex-concave constant-sum games, we rely on online convex optimization where each player minimizes its own convex loss. See for example \citep{PLG,nesterov2005excessive,hoda2010smoothing}. {\em Regret minimization} has been extensively considered in games since the average strategy of self-playing no-regret algorithms converges to a Nash equilibrium~\citep{rakhlin2013optimization,kangarshahi2018let}. Counterfactual regret minimization (CFR) has been considered in the setting of imperfect information games in \citep{CFR} showing a $O(1/\sqrt{t})$ convergence rate in terms of exploitability. Other techniques provide a faster rate of convergence $O(1/t)$ \citep{daskalakis2011near,syrgkanis2015fast,abernethy2018faster,Farina19:Optimistic}. These techniques have been usually studied in the discrete time setting but has also been looked at in continuous time~\citep{mertikopoulos2018cycles}.

\paragraph{Convergence of the last iterate.} 
{\em Extragradient or optimistic mirror descent} methods have been proven to converge to a Nash equilibrium~\citep{korpelevich1976extragradient,MertikopoulosLZ19} with possibly an exponential rate in unconstrained spaces~\citep{mokhtari2019unified}. The most closely related extragradient method in this domain is optimistic multiplicative-weights-update \citep[OMWU; ][]{daskalakis2018last} which provides convergence guarantees to the Nash equilibrium of the last iterate.
Another approach uses the Frank-Wolfe method to compute Nash equilibria in normal-form games \citep{gidel2016frank}, although convergence is attained at the same rate as for fictitious play. 
A related algorithm introduced by \citet{MAIO} for imperfect information games consists in each player doing a step of {\em mirror ascent against an improved opponent} (MAIO) for which exponential convergence of the last-iterate was proven (with a instance-dependent exponent). 
Other works include the {\em regularized Nash dynamics}  \citep{Perolat2021,Perolat_2022} under continuous-time dynamics, the Magnet Mirror Descent \cite{sokota2023unified} which are related to Online Mirror Descent (OMD) performed on the regularized game, and the MTPO algorithm of  \citet{shani2024multiturn} in the context of multi-turn LLMs. A thorough comparison between Nash-MD and OMD is given in the next section.

\section{Analysis of a tabular algorithm: Nash-MD}\label{sec:Nash-MD}

For simplicity of notation we remove the dependence on the context $x$, thus policies $\pi\in \Delta(\actionspace)$ are probability distributions over $\actionspace$. We now introduce an algorithm, called {\bf Nash-MD}, which is a novel variant of mirror descent  \citep{Nemirovski1983,Bubeck15,Lattimore2020} that makes use of a specific regularized policy $\pi_t^\mu$ which is a geometric mixture between the current policy $\pi_t$ and the reference policy $\mu$. We prove the convergence (in terms of \KL~distance) of the last iterate to the Nash equilibrium of $\P_\tau$.

\paragraph{The Nash-MD algorithm:}
Define the regularized policy $\pi_t^\mu$ as a geometric mixture between the current policy $\pi_t$ and the reference policy $\mu$: 
\beq\label{eq:geometric-mixture}
\pi_t^\mu(y) \eqdef \frac{\pi_t(y)^{1-\eta_t\tau}\mu(y)^{\eta_t\tau}}{\sum_{y'}\pi_t(y')^{1-\eta_t\tau}\mu(y')^{\eta_t\tau}},
\eeq
where $\eta_t$ is a learning rate. We define the {\bf Nash-MD algorithm} as a step of mirror descent relative to the regularized policy $\pi_t^\mu$:
\beq\label{eq:Nash.MD}
\pi_{t+1}\eqdef \arg\max_\pi \left[ \eta_t \P(\pi\succ \pi_t^\mu) - \KL(\pi, \pi_t^\mu)\right].
\eeq
The optimization above can also be made explicit in the following form:
\beqan
\pi_{t+1}(y)\propto\pi_t^\mu(y)\exp\left(\eta_t \P(y\succ \pi_t^\mu)\right),
\eeqan
or equivalently
\beqa\label{eq:Nash.MD.log}
\log \pi_{t+1}(y) &=&  (1-\eta_t\tau)\log \pi_t(y) + \eta_t\tau\log \mu(y) \notag \\
& &+ \eta_t \P(y\succ \pi_t^\mu) + c,
\eeqa
where $c$ is a normalization constant that is independent of $y$.

The intuition for this algorithm is to improve the current policy $\pi_t$ in a direction that increases the preference $\pi\mapsto\P(\pi, \pi_t^\mu)$ against the regularized policy $\pi_t^\mu$, while not deviating too much from it.  
\newpage
We now state our main theoretical result; see Appendix~\ref{sec:proof.Theorem} for the proof.

\begin{theorem}\label{thm:MD}
Let $\pi^*_\tau$ be the Nash equilibrium of the regularized preference model:
$\P_\tau(\pi\succ \pi') = \P(\pi\succ \pi') - \tau \KL(\pi, \mu) +\tau \KL(\pi', \mu).$ At every iteration $t$ we have that
\beq\label{eq:nash.md}
\KL(\pi^*_\tau,\pi_{t+1}) \leq (1-\eta_t\tau) \KL(\pi^*_\tau, \pi_t) + 2 \eta_t^2.
\eeq
We deduce that for the choice $\eta_t=2/(\tau (t+2))$ we have $$\KL(\pi^*_\tau,\pi_T) \leq  \frac{8}{\tau^2(T+1)}.$$ 
\end{theorem}

Thus this algorithm produces a sequence of policies $(\pi_t)_{1\leq t\leq T}$ with last-iterate convergence (in \KL-divergence) to the regularized Nash equilibrium $\pi^*_\tau$ at a speed $O(1/T)$. We now mention several important features of this algorithm, specially in the context of LLMs.

\paragraph{Nash-MD does not require playing against the full mixture $\bar\pi_t$.}
In order to compute $\pi_{t+1}$ we do not need to play against the mixture $\bar\pi_t=\frac 1t\sum_{s=1}^t\pi_s$ of past policies (where by `playing against a policy $\tilde{\pi}$' we mean computing (or estimating) the preference $\P(y\succ \tilde{\pi})$), unlike in fictitious play. We play against a single (geometric) mixture $\pi_t^{\mu}$ between the current policy $\pi_t$ and the reference policy $\mu$. This is important in situations, such as in LLMs, where storing and generating sample from several policies is costly.

\paragraph{Nash-MD has a last-iterate convergence property.}
The second important property of Nash-MD is that we have convergence of the last-iterate (i.e., the current policy $\pi_t$ converges to $\pi_\tau^*$) and not only convergence on average (as is typically the case of fictitious play and usual regret minimization algorithms like CFR and OMD). 
This feature is particularly important in the context of LLMs as well due to the substantial memory resources that would be otherwise needed to store a mixture policy like $\bar\pi_t$.

\paragraph{Comparison with online mirror descent (OMD).}
In general the analysis of constant-sum concave-convex games can be performed in the framework of online convex optimization where the goal is to find a sequence of solutions $\pi_t$ that minimizes the sum of a sequence of convex loss functions $\pi\mapsto l_t(\pi)$. The OMD algorithm (using the KL as Bregman divergence) defines the sequence:
\beq\label{eq:omd}
\pi_{t+1} \eqdef \arg\min_\pi \left[\eta_t \nabla l_t(\pi_t) \cdot(\pi-\pi_t) + \KL(\pi,\pi_t)\right],
\eeq
for which it can be shown (see \citet{PLG}) that the average cumulative regret, under optimal choice of learning rate, can be bounded as
{\small $$\frac{1}{T} \sum_{t=1}^T l_t(\pi_t) - \min_\pi \frac{1}{T} \sum_{t=1}^T l_t(\pi) = O\left(1/\sqrt{T}\right).$$ }
This type of upper bound on the regret can be further used to obtain convergence of constant-sum games where each player follows an OMD strategy to minimize their own convex loss. In our context, we could apply this OMD strategy to minimize the regularized preference model $\P_\tau$, and since $\P_\tau$ is antisymmetric, we only need to consider the dynamics of a single player. 
So the loss function at time $t$ is the negative preference against the current policy of the opponent: $l_t(\pi)=-\P_\tau(\pi\succ \pi_t)$. 
We deduce that 

$\nabla l_t(\pi_t) =- \left[ \partial_\pi \P_\tau(\pi\succ \pi_t)\right]_{\pi=\pi_t}$, thus 

$\nabla l_t(\pi_t)\cdot \pi = - \sum_y \pi(y) \left[\P(y\succ \pi_t) -\tau\left(\log\frac{\pi_t(y)}{\mu(y)}+1\right)\right]$. 

The OMD update rule in \eqref{eq:omd} can be rewritten as
{\small
\beqan
\pi_{t+1}&=&\arg\max_\pi \Big[ \eta_t \sum_y \pi(y) \Big(\P(y\succ \pi_t)-\tau\log\frac{\pi_t(y)}{\mu(y)}\Big) \\
& & - \KL(\pi, \pi_t)\Big].
\eeqan
}
Now, using the regularized policy $\pi_t^\mu$ introduced in \eqref{eq:geometric-mixture}, we can rewrite this update rule as
\beq\label{eq:OMD.short}
\pi_{t+1}=\arg\max_\pi \left[ \eta_t \P(\pi\succ \pi_t) - \KL(\pi, \pi_t^\mu)\right].
\eeq
Comparing \eqref{eq:Nash.MD} and \eqref{eq:OMD.short} we notice that both OMD and Nash-MD make use of the same KL penalty term $\KL(\pi,\pi_t^\mu)$. However they differ in the fact that OMD optimizes the preference $\pi\mapsto \P(\pi\succ\pi_t)$ against the current policy $\pi_t$ whereas Nash-MD optimizes the preference $\pi\mapsto \P(\pi\succ \pi_t^\mu)$ against the regularized policy $\pi_t^\mu$.

In the context of convex-concave games, the regret bound on the average cumulative regret translates into an upper bound on the exploitability
of the game when players play their average policies, thus entailing their on-average convergence to the Nash equilibrium. However it is known that usual regret-minimization algorithms may not possess a last-iterate convergence property because the sequence of policies $\pi_t$ may oscillate around the Nash equilibrium (see, for example, \citealp{mertikopoulos2018cycles}). Nevertheless, last-iterate convergence have been obtained for variants of OMD, such as extra-gradient and optimistic versions, see e.g., \citep{rakhlin2013optimization,daskalakis2018last,MertikopoulosLZ19,MAIO,mokhtari2019unified}. In the context of LLMs, the MTPO algorithm of \citet{shani2024multiturn}, defined by the same update rule (\ref{eq:OMD.short}), is shown to converge, in the last-iterate, to the regularized Nash equilibrium, but with a speed that depends on the inverse minimum non-zero probability of the reference policy $\piref_{\min}=\min_{y:\piref(y)>0} \piref(y)$. Also the on-line version of IPO \citep{calandriello2024human} offers a deep-learning version of OMD similar to the update rule defined by \eqref{eq:OMD.short}.

To the best of our knowledge, it appears that Nash-MD has not been introduced before, despite its simplicity. Nash-MD enjoys a last-iterate convergence property with a KL-divergence to the Nash equilibrium decaying as $O(1/T)$ (with a constant in the big $O$ notation independent of $\piref_{\min}$). We believe the reason this simple modification of OMD possesses these nice properties is because of the special structure of the regularized preference function that we consider here which is the sum of a bilinear function (in policy spaces) and a KL-penalty term.

\paragraph{Comparison with Ruppert-Polyak averaging.}
The weighted averaging of log-probabilities in \eqref{eq:Nash.MD.log}, along with the updates in OMD and other algorithms for regret minimisation and learning in games \citep{PLG}, also bear relation with Ruppert-Polyak averaging \citep{ruppert1988efficient,polyak1990new,polyak1992acceleration}, a general technique in stochastic approximation in which iterates are averaged to accelerate convergence. However, the specific form of averaging and its uses within Nash-MD (over log-probabilities, and for use in opponent policies) are essential for shaping the dynamics of the algorithm and establishing convergence, not just as a method for acceleration.

\paragraph{The contextual bandit setting.}
All the results mentioned in this section are for the state-independent case, where policies and preferences do not depend on the context $x$. In the case of LLMs the context is the prompt $x$, and responses $y$ and $y'$ are generated conditioned on $x$. However the theoretical results do not change. Indeed, we would define the Nash-MD algorithm in the contextual bandit case as follows: for every $x\in{\tt supp}(\rho)$,
\beqan
\pi_{t+1}(\cdot|x)&\eqdef& \arg\max_{\pi(\cdot)} \big[ \eta_t \P(\pi(\cdot|x)\succ \pi_t^\mu(\cdot|x)|x) \\
& & - \KL(\pi(\cdot), \pi_t^\mu(\cdot|x))\big],
\eeqan
where
$$\pi_t^\mu(y|x) \propto \pi_t(y|x)^{1-\eta_t\tau}\mu(y|x)^{\eta_t\tau}.$$
We prove the convergence of this algorithm, in exactly the same way as in Theorem~\ref{thm:MD}, by showing that at every iteration $t$ we have 
\begin{eqnarray*} 
\KL(\pi^*_\tau,\pi_{t+1}) &\leq& (1-\eta_t\tau) \KL(\pi^*_\tau, \pi_t) + 2 \eta_t^2,
\end{eqnarray*}
where $\KL(\pi,\pi')=\E_{x\sim\rho}[\KL(\pi(\cdot|x),\pi'(\cdot|x))]$.

\begin{table*}[ht]
\caption{PaLM 2 preference $\P^*(\pi_c\succ\pi_r)$ model between column policy $\pi_c$ against row policy $\pi_r$.}
\label{results:p*.text}
\centering
{\small
\setlength\tabcolsep{4.7pt}
\begin{tabular}{c|l|l|l|l|l|l|l|l|l|l|l|l|l|l}
$\P^*$ &   SFT &  RLHF &    SP &   {\bf MD1} &   MD2 &   MD3 &   MD4 &   MD5 &   MD6 &    BR &  EMA1 &  EMA2 &  EMA1* &  EMA2* \\
\midrule
  SFT & 0.500 & 0.990 & 0.983 & {\bf 0.982} & 0.989 & 0.987 & 0.985 & 0.982 & 0.965 & 0.943 & 0.970 & 0.961 &  0.977 &  0.980 \\ \cline{4-11}
  RLHF & 0.010 & 0.500 & \multicolumn{1}{l|}{{\color{blue}{\bf 0.489}}} &
  \multicolumn{1}{l|}{{\color{blue}{\bf 0.598}}} &
  \multicolumn{1}{l|}{{\color{blue}{\bf 0.519}}} &
  \multicolumn{1}{l|}{{\color{blue}{\bf 0.561}}} & \multicolumn{1}{l|}{{\color{blue}{\bf 0.501}}} & \multicolumn{1}{l|}{{\color{blue}{\bf 0.436}}} & \multicolumn{1}{l|}{{\color{blue}{\bf 0.284}}} &
  \multicolumn{1}{l|}{{\color{blue}{\bf 0.148}}} & 
  0.468 & 0.320 &  0.477 &  0.510 \\ \cline{4-11}
    SP & 0.017 & 0.511 & 0.500 & {\bf 0.592} & 0.504 & 0.545 & 0.499 & 0.451 & 0.310 & 0.211 & 0.445 & 0.362 &  0.464 &  0.488 \\
   MD1 & 0.018 & 0.402 & 0.408 & {\bf 0.500} & 0.425 & 0.470 & 0.369 & 0.362 & 0.238 & 0.163 & 0.391 & 0.270 &  0.400 &  0.447 \\
   MD2 & 0.011 & 0.481 & 0.496 & {\bf 0.575} & 0.500 & 0.513 & 0.491 & 0.434 & 0.298 & 0.196 & 0.460 & 0.351 &  0.430 &  0.496 \\
   MD3 & 0.013 & 0.439 & 0.455 & {\bf 0.530} & 0.487 & 0.500 & 0.484 & 0.408 & 0.273 & 0.187 & 0.429 & 0.323 &  0.413 &  0.472 \\
   MD4 & 0.015 & 0.499 & 0.501 & {\bf 0.631} & 0.509 & 0.516 & 0.500 & 0.428 & 0.265 & 0.161 & 0.468 & 0.358 &  0.437 &  0.503 \\
   MD5 & 0.018 & 0.564 & 0.549 & {\bf 0.638} & 0.566 & 0.592 & 0.572 & 0.500 & 0.329 & 0.210 & 0.532 & 0.389 &  0.518 &  0.539 \\
   MD6 & 0.035 & 0.716 & 0.690 & {\bf 0.762} & 0.702 & 0.727 & 0.735 & 0.671 & 0.500 & 0.342 & 0.652 & 0.548 &  0.651 &  0.691 \\
    BR & 0.057 & 0.852 & 0.789 & {\bf 0.837} & 0.804 & 0.813 & 0.839 & 0.790 & 0.658 & 0.500 & 0.743 & 0.640 &  0.752 &  0.774 \\
  EMA1 & 0.030 & 0.532 & 0.555 & {\bf 0.609} & 0.540 & 0.571 & 0.532 & 0.468 & 0.348 & 0.257 & 0.500 & 0.381 &  0.480 &  0.556 \\
  EMA2 & 0.039 & 0.680 & 0.638 & {\bf 0.730} & 0.649 & 0.677 & 0.642 & 0.611 & 0.452 & 0.360 & 0.619 & 0.500 &  0.585 &  0.659 \\
 EMA1* & 0.023 & 0.523 & 0.536 & {\bf 0.600} & 0.570 & 0.587 & 0.563 & 0.482 & 0.349 & 0.248 & 0.520 & 0.415 &  0.500 &  0.555 \\
 EMA2* & 0.020 & 0.490 & 0.512 & {\bf 0.553} & 0.504 & 0.528 & 0.497 & 0.461 & 0.309 & 0.226 & 0.444 & 0.341 &  0.445 &  0.500 \\
\end{tabular}
}
\end{table*}

\section{Deep learning implementation of NLHF}
We build upon the insights from Nash-MD and describe gradient-based algorithms {\bf Nash-MD-PG} and {\bf Nash-EMA-PG} for deep-learning architectures designed for the computation of the Nash equilibrium of a preference model, with a specific focus on their applicability in the context of LLMs. The general form of the policy gradient is 
$$
\nabla_\theta \log\pi_\theta(y|x) \left( \P(y\succ y'|x)-1/2 -\tau \log\frac{\pi_\theta(y|x)}{\mu(y|x)}\right),
$$
 where the prompt $x\sim\rho$, the first responses $y\sim\pi_\theta(\cdot|x)$ and the second response $y'\sim \pi'(\cdot|x)$ where the {\em alternative policy $\pi'$} is either
 \begin{itemize}
     \item the (geometric) mixture policy $\pi'(y|x)\propto (\pi_\theta(y|x))^{1-\beta}(\mu(y|x))^\beta$ between $\pi_\theta$ and $\mu$ (for some mixture parameter $\beta\in[0,1]$), in the case of the {\bf Nash-MD-PG} algorithm,
     \item the policy with a parameter being an exponentially moving average of past parameters, in the case of the {\bf Nash-EMA-PG} algorithm. 
 \end{itemize} 

Notice we have subtracted the baseline $1/2 = \P(y\succ y|x)$ from the preference $\P(y\succ y'|x)$ (which does not change the expectation of the gradient) as a variance reduction technique that does not require learning a value function. All the details of these algorithms are given in Appendix~\ref{sec:implementations}.

\section{Experiments on a text summarization task}
In Appendix~\ref{apx:experiments} we report experiments on a text summarization task and compare several algorithms for NLHF (Self-Play, Best-Response against $\mu$, Nash-MD-PG and Nash-EMA-PG) as well as a RLHF baseline.

We made a pairwise evaluation of all the models by querying a very large LLM (PaLM 2 Large) \citep{anil2023palm} to obtain a preference signal, which is reported in Table~\ref{results:p*.text}. The models that are compared are respectively: SFT (Supervised Fined-Tuned, from which all other models are initialized and regularized to, i.e., this is also $\piref$), RLHF, SP (Self-Play, equivalent to Nash-MD-PG with $\beta=0$), MD1-MD6 (Nash-MD-PG with $\beta\in \{0.125, 0.25, 0.375, 0.5, 0.625, 0.75\}$), BR (Best Response against SFT, equivalent to Nash-MD-PG with $\beta=1$), EMA1-2 (last-iterate of Nash-EMA-PG with $\beta\in\{0.999, 0.9995\}$), and EMA$1^*-2^*$ (average policy of Nash-EMA-PG).
See the full description in Appendix~\ref{apx:experiments}.

The Nash-MD-PG models (specially for $\beta\in[0.125,0.375]$) emerge as the best-performing method, surpassing the other models in this pairwise comparison.

The choice of the mixture parameter $\beta$ in Nash-MD-PG entails an interesting trade-off (see the numbers highlighted in blue). A parameter value $\beta=0$ corresponds to Self-Play, while a value $\beta=1$ represents Best-Response against the initial policy (SFT). Notably, intermediate values within the range of 0.125 to 0.375 consistently outperform both Self-Play and Best-Response, highlighting the advantages of self-improvement when playing against a mixture policy (between self and a past version of self) as opposed to against a pure policy (either self or fixed).

Notice that it is difficult to establish a fair comparison between the NLHF and RLHF approaches since they rely on different models: a preference model for NLHF versus a reward model for RLHF, and the quality of the learnt models should be part of the picture for a full comparison between the NLHF and RLHF approaches. Thus the goal of these experiments is not to show the superiority of a method over another one (this would also  require a more intensive and larger scale empirical evaluation) but rather to illustrate how the proposed NLHF approach, and in particular the Nash-MD algorithm, can be implemented in a practical LLM setting.


This is also the reason we have not tried to over-optimize the hyper-parameters (such as the learning rate, the number of learning steps, etc) of the different methods (and we used the same parameters for all NLHF algorithms) and this could explains some differences observed in the pairwise preference Table compared with the results reported in \citep{calandriello2024human}, in which they optimize each algorithm with a different set of hyper-parameters.

\section{Conclusion and future work}

NLHF emerges as an interesting and promising alternative to RLHF, offering a fresh perspective on aligning models with human preferences. Given human pairwise preference data, learning a preference model is a more intuitive and natural approach compared to learning a reward model. It involves simpler techniques, such as supervised learning, and doesn't requires making specific assumptions, such as Bradley-Terry.
Once a preference model is established, the concept of the Nash equilibrium naturally arises as a compelling solution concept. 

For this new NLHF framework we have introduced Nash-MD, an algorithm that optimizes policies by following a self-improvement technique where the current model improves itself by playing (i.e., by generating and comparing responses) against a (geometric) mixture of the current model and a past model. The parameter $\beta$ of the mixture ranges from $\beta=0$ (which corresponds to playing against itself) to $\beta=1$ (playing against a fixed policy) and may take any value in between. 

In the case of tabular policy representations we have establish its last-iterate convergence to the Nash equilibrium of the regularized preference model.
We have also introduced and implemented deep learning versions Nash-MD-PG and Nash-EMA-PG inspired by Nash-MD, and described how these ideas can be applied to LLMs by reported experimental results on a text-summarizing task.

Future research directions would consider the exploration of various mixtures between the current policy and past checkpoints, extending the concept initially introduced by Nash-MD. Additionally, another immediate direction would consider incorporating a decaying mixing coefficient $\beta\rightarrow 0$ to the deep Nash-MD variants to align more closely with theoretical considerations.

In conclusion, NLHF offers a compelling avenue for preference learning and policy optimization for aligning models with human preferences. An an example of a possible NLHF implementation we have introduced Nash-MD as a possible algorithmic solution and described some theoretical properties as well as deep learning adaptations. Further research in this direction, including the use of different mixture strategies, holds significant promise for advancing the field of aligning LLMs with human preferences.

\clearpage

\section*{Impact statement}
This paper presents work whose goal is to advance the field of Machine Learning. There are many potential societal consequences of our work, none which we feel must be specifically highlighted here.

\section*{Acknowledgements}
We would like to thank the individuals who designed and built the RL training infrastructure used in this paper: Eugene Tarassov, Léonard Hussenot, Johan Ferret, Robert Dadashi, Geoffrey Cideron, Alexis Jacq, Sabela Ramos, Piotr Stanczyk, Danila Sinopalnikov, Amélie Héliou, Ruba Haroun, Matt Hoffman, Bobak Shahriari, and in particular Olivier Pietquin for motivating discussions. We would like to express our gratitude to Ivo Danihelka, David Silver, Guillaume Desjardins, Tor Lattimore, and Csaba Szepesv\'ari for their feedback on this work. Finally we would like to thank the anonymous reviewers who helped us improve the quality of this final version.

\bibliographystyle{icml2024}
\bibliography{biblio}

\newpage

\appendix
\onecolumn
\section{Maximizing expected Elo vs maximizing probability of winning}\label{sec:Elo.vs.winning.proba}
Consider the following preference model, where the set of actions is $\actionspace=\{y_1,y_2,y_3\}$ and the preference table between these actions is

\begin{center}
\begin{tabular}{c|c|c|c}
 $\P(y\succ y')$   & $y=y_1$ & $y=y_2$ & $y=y_3$ \\
\midrule $y'=y_1$ &  1/2  &  9/10 & 2/3   \\
\hline $y'=y_2$ &  1/10 &  1/2  & 2/11  \\
\hline $y'=y_3$ &  1/3  &  9/11 &  1/2  \\
\end{tabular}
\end{center}

This preference model can be perfectly captured by a Bradley-Terry reward model in which the Elo score of the actions would be (up to an additive constant): $R(y_1)= 0$, $R(y_2)=\log 9$, and $R(y_3)=\log 2$. 

If we optimize over the simplex $\Delta(\actionspace)$, then the policy selecting deterministically $y_2$ is optimal both in terms of rewards and in terms of preference against any policy. However, if we consider a constrained optimization problem where we search for a policy in a subset ${\cal S}\subset \Delta(\actionspace)$, then the optimum of the expected reward and preference may be different. To illustrate, let ${\cal S}$ be the set of probability distributions $\pi\in\Delta(\actionspace)$ such that $\pi(y_1)=2\pi(y_2)$.

In that case, the policy $\pi^*_{R}\eqdef (2/3, 1/3, 0)$ is optimal in terms of maximizing expected rewards whereas the policy  $\pi^*_{\P}\eqdef (0, 0, 1)$  is optimal in terms of maximizing preference against any alternative policy in ${\cal S}$. 
In particular we have 
$$\E_{y\sim \pi^*_{R}}[R(y)] = 0\times 2/3 + \log(9)\times 1/3 > \log(2) = \E_{y\sim \pi^*_{\P}}[R(y)],$$
whereas policy $\pi^*_{\P}$ is preferred over $\pi^*_{R}$, since
$$\P(\pi^*_{\P}\succ \pi^*_{R}) = \P(y_3\succ y_1) \times 2/3 + \P(y_3\succ y_2) \times 1/3 = 50/99 > 1/2.$$

Thus if one searches for a policy in ${\cal S}$, then the optimum in terms of maximizing expected (Elo) reward and maximizing preference (probability of winning) are different.

Note that the constraint $\pi\in {\cal S}$ may be imposed in a soft way using regularization. Here for example we could implement a 2-step decisions process where in a first step one would choose the probability mass assigned to $y_3$, and in the second step, one would choose the remaining mass to allocate between $y_1$ and $y_2$. 
The second step may be constrained in a soft way by penalizing distributions (over $y_1$ and $y_2$) that are different from a reference distribution $\mu=(2/3,1/3)$ by using a KL-regularization with a large $\tau$ coefficient. In this way the set of effective policies that would be considered would be close to ${\cal S}$.

This example illustrates the fact that in constrained (or regularized) optimization settings, maximizing Elo versus preference are different objectives, even in a setting where preferences can be perfectly expressed in a Bradley-Terry model. 

\section{Sensitivity of  reward models w.r.t.~the sampling distribution}\label{sec:data.dependent.reward.models}

\begin{proposition}\label{prop:best.BT}
For a given preference model $\P(y\succ y')$ and a distribution $\pi$, let us define the best Bradley-Terry reward model:
\beq\label{eqn:best.BT.model}
r^{\pi}\eqdef\arg\max_{r}
\E_{
{\tiny
\begin{array}{c}
y,y'\sim \pi(\cdot) \\
Z\sim\nu
\end{array}
}}
\left[ \log\left(\sigma(r(y_w^Z)-r(y_l^Z))\right)\right]
\hspace{-3mm}
\eeq
where $y_w^Z$ and $y_l^Z$ are respectively the preferred (and less preferred) response (among $y$ and $y'$ sampled from $\pi(\cdot)$) according to a randomly sampled human $Z\sim\nu$.

Define the corresponding Bradley-Terry preference model $\P_{BT}^\pi(y\succ y')\eqdef \sigma(r^\pi(y) - r^\pi(y'))$. Then we have that for any $y$ in the support of $\pi$,
$$\P(y\succ \pi) = \P_{BT}^\pi(y\succ \pi).$$
\end{proposition}

\begin{proof}
First notice that, from the definition of $y_w^Z$ and $y_l^Z$, we have
$$r^{\pi}=\arg\max_{r} {\mathbb E}_{y,y'\sim\pi} \left[\P(y\succ y') \log\sigma(r(y)-r(y'))\right].
$$
Write ${\cal L}(r)$ the loss that is minimized. Thus $r^\pi=\arg\max_r {\cal L}(r)$, and
\begin{eqnarray*}
\nabla {\cal L}(r)&=&{\mathbb E}_{y,y'\sim\pi} \left[\P(y\succ y') \frac{\nabla \sigma(r(y)-r(y'))}{\sigma(r(y)-r(y'))}\right]\\
&=& {\mathbb E}_{y,y'\sim\pi} \left[\P(y\succ y') \sigma(r(y')-r(y)) (\nabla r(y)-\nabla r(y'))\right]
\end{eqnarray*}
Thus looking a the $z$-th element of $\nabla {\cal L}(r^\pi)$:
\begin{eqnarray*}
\partial_{r(z)} {\cal L}(r^\pi)&=& \pi(z){\mathbb E}_{y\sim\pi} \left[\P(z\succ y) \sigma(r^\pi(y)-r^\pi(z))  - (1-\P(z\succ y)(1-\sigma(r^\pi(y)-r^\pi(z)))\right] \\
&=& \pi(z){\mathbb E}_{y\sim\pi} \left[\P(z\succ y) + \P_{BT}^\pi(y\succ z) - 1\right] \\
&=& \pi(z)\left[\P(z\succ \pi) - \P_{BT}^\pi(z\succ \pi)\right].
\end{eqnarray*}
Setting the derivative $\partial_{r(z)} {\cal L}(r^\pi)=0$ we deduce that for $z$ in the support of $\pi$ we have $\P(z\succ \pi) = \P_{BT}^\pi(z\succ \pi)$.
\end{proof}

\begin{theorem}[The optimal BT-reward model depends on the sampling distribution] \label{thm:data.dependent.reward.models}
If a preference model $\P$ cannot be perfectly captured by a Bradley-Terry reward model, in the sense that the preference model $\P_{BT}^\pi(y\succ y')\eqdef \sigma\left( r^{\pi}(y) - r^{\pi}(y') \right)$ corresponding to the best Bradley-Terry reward model $r^\pi$, solution to \eqref{eqn:best.BT.model} for some policy $\pi$, is not identical to $\P$, then the reward model $r^\pi$ depends explicitly on the sampling distribution $\pi$. 

More precisely, if there exists $y,y'$ and $\pi$ such that $\P_{BT}^\pi(y\succ y')\neq \P(y\succ y')$, then there exists another policy $\pi'\neq\pi$ (with same support as $\pi$) such that $r^\pi(y) - r^\pi(y') \neq r^{\pi'}(y) - r^{\pi'}(y')$. Thus we also have that $\P_{BT}^\pi(y\succ y')\neq \P_{BT}^{\pi'}(y\succ y')$.
\end{theorem}
This result shows that the reward model $r^\pi$ (thus the corresponding Bradley-Terry preference model $\P_{BT}^\pi$) depends on the sampling distribution $\pi$ (which explains our use of $\pi$ as superscript).

\begin{proof}
Assume there exists $y,y'$ and $\pi$ such that $\P_{BT}^\pi(y\succ y')\neq \P(y\succ y')$. Let us define $\pi'$ (with same support as $\pi$) as follows:
$\pi'(z)=\frac 12 \pi(z)$ for $z\neq y'$, and $\pi'(y')=c \pi(y')$ for some constant $c>1$ (defined such that $\pi'$ is a proper probability distribution). We deduce that
$$
{\cal Q}(y\succ\pi') = \frac 12 {\cal Q}(y\succ\pi) +(c-1/2)\pi(y'){\cal Q}(y\succ y'),
$$
for any preference model ${\cal Q}$. Applying this equality both with ${\cal Q}=\P$ and ${\cal Q}=\P_{BT}^\pi$, and since $\P_{BT}^\pi(y\succ y')\neq \P(y\succ y')$ and $\P_{BT}^\pi(y\succ \pi)= \P(y\succ \pi)$ (from Proposition~\ref{prop:best.BT}), we deduce that $\P_{BT}^\pi(y\succ \pi')\neq \P(y\succ \pi')$. Applying Proposition~\ref{prop:best.BT} again we have that $\P(y\succ \pi')=\P_{BT}^{\pi'}(y\succ \pi')$, thus
$$\P_{BT}^\pi(y\succ \pi')\neq \P_{BT}^{\pi'}(y\succ \pi').$$
We deduce that 
$$\sum_{z} \pi'(z)\left[ \sigma\big( r^\pi(y) - r^\pi(z)\big) - \sigma\big( r^{\pi'}(y) - r^{\pi'}(z)\big) \right] \neq 0,$$
thus there exists (at lease one) $z$ such that  $r^\pi(y) - r^\pi(z) \neq r^{\pi'}(y) - r^{\pi'}(z).$
This concludes the proof that the two BT-reward models $r^\pi$ and $r^{\pi'}$ are different as well as the corresponding BT-preference models $\P_{BT}^\pi$ and $\P_{BT}^{\pi'}$.
\end{proof}

\section{Preference models may be non-transitive}\label{apx:non-transitive}
\subsection{Example of a non-transitive preference model}
Notice that in general a preference model may not be transitive. Here is a simple illustration of a non-transitive preference model where we have 3 policies $\pi_1$, $\pi_2$ and $\pi_3$ such that $\P(\pi_1\succ\pi_2)>1/2$, $\P(\pi_2\succ\pi_3)>1/2$ and $\P(\pi_3\succ\pi_1)>1/2$. 

We consider the set of outcomes being the subset of integers $\actionspace=\{1, 2, \dots, 9\}$ and the 3 policies being defined by $\pi_1 = {\cal U}(\{2, 4, 9\})$, $\pi_2 = {\cal U}(\{1, 6, 8\})$, and $\pi_3 = {\cal U}(\{3, 5, 7\})$, where ${\cal U}(S)$ refers to a uniform distribution over the set $S$. The preference is defined as
${\cal P}(\pi\succ\pi')=\E_{y\sim \pi, y'\sim\pi'} [\1{y \geq y'}]$. Then we have 
$$ {\cal P}(\pi_1\succ\pi_2) = {\cal P}(\pi_2\succ\pi_3) = {\cal P}(\pi_3\succ\pi_1) = 5/9.$$ 
This mirrors the classical example of non-transitive dice \citep{gardner1970paradox}.

\subsection{Non-transitive aggregation of individual transitive preferences}\label{apx:non-transitive.2}

Here we show that even if each human has transitive individual preferences, the resulting average preference model may not be transitive. Let us consider a specific case of a preference model defined as the probability (under some random outcome $\randomoutcome$) that $f(x,y,\randomoutcome)\geq f(x,y',\randomoutcome)$, where $f$ is a (deterministic) absolute scoring function:
$${\cal P}(y\succ y'|x) = \E_{\randomoutcome\sim\randomoutcomedist} \left[ \1{f(x,y,\randomoutcome)\succ f(x,y',\randomoutcome)}\right],$$
where we define the function $\1{u\succ v}\eqdef (\sign(u-v)+1)/2$, which behaves as an indicator for the event $u > v$, and assigning a value of $1/2$ in the case where $u=v$. For example, this could represent the probability that a randomly selected human $\randomoutcome\sim\randomoutcomedist$ prefers choice $y$ over choice $y'$ in a context $x$.

Consider the following example, where there are 3 possible responses $y_1, y_2, y_3$ and 3 possible humans $z_1, z_2,z_3$ chosen uniformly at random: $\nu={\mathcal U}(\{z_1, z_2,z_3\})$. Define the scoring function $f$ as follows:
\begin{eqnarray*}
&f(y_1,z_1)= 2, \quad f(y_1,z_2)= 4, \quad f(y_1,z_3)=9, \\
&f(y_2,z_1)= 1, \quad f(y_2,z_2)= 6, \quad f(y_2,z_3)=8, \\
&f(y_3,z_1)= 3, \quad f(y_3,z_2)= 5, \quad f(y_3,z_3)=7.
\end{eqnarray*}

Notice that this defines a transitive preference model for each individual human $z\in\{z_1,z_2,z_3\}$. However, when aggregated the preference model satisfies ${\cal P}(y_1\succ y_2) = {\cal P}(y_2\succ y_3) = {\cal P}(y_3\succ y_1) = 2/3$. This example thus illustrates that even if for each individual, preferences are totally ordered, when averaged over humans, the resulting preferences model may be non-transitive.

\section{Proof of Theorem \ref{thm:MD}}\label{sec:proof.Theorem}
We start with a first lemma.
\begin{lemma}\label{lem:mixture}
For any $\pi$, and $0\leq \eta_t\tau\leq 1$, we have
$$\KL(\pi , \pi_t^\mu) \leq  \eta_t\tau \KL(\pi , \mu) + (1-\eta_t\tau)\KL(\pi , \pi_{t}) - \eta_t\tau \KL(\pi_t^\mu, \mu).$$
\end{lemma}
\begin{proof}
From the definition of $\pi_t^\mu$, we have 
$$\log \pi_t^\mu(y)= (1-\eta_t\tau) \log \pi_{t}(y) + \eta_t\tau \log \mu(y) - \log Z, $$
where we define $Z = \sum_{y'} (\pi_{t}(y'))^{1-\eta_t\tau} (\mu(y'))^{\eta_t\tau}$. Thus, for any $\pi$, we have
$$\KL(\pi , \pi_t^\mu) = \eta_t\tau \KL(\pi , \mu) + (1-\eta_t\tau)\KL(\pi , \pi_{t}) + \log Z.$$
We have that
\begin{eqnarray*}
\eta_t\tau \KL(\pi_t^\mu, \mu) &=& \eta_t\tau \sum_y \pi_t^\mu(y) \log\frac{\big(\pi_{t}(y)\big)^{1-\eta_t\tau}\big(\mu(y)\big)^{\eta_t\tau}}{Z \mu(y)} \\
&=&(1-\eta_t\tau) \sum_y \pi_t^\mu(y) \log\frac{\big(\pi_{t}(y)\big)^{\eta_t\tau}}{ \big(\mu(y)\big)^{\eta_t\tau}} - \eta_t\tau \log Z\\
&\leq &(1-\eta_t\tau) \log \sum_y \pi_t^\mu(y) \frac{\big(\pi_{t}(y)\big)^{\eta_t\tau}}{ \big(\mu(y)\big)^{\eta_t\tau}} - \eta_t\tau \log Z\\
&=& (1-\eta_t\tau) \log \sum_y \frac{\big(\pi_{t}(y)\big)^{1-\eta_t\tau}\big(\mu(y)\big)^{\eta_t\tau}}{Z} \frac{\big(\pi_{t}(y)\big)^{\eta_t\tau}}{ \big(\mu(y)\big)^{\eta_t\tau}} - \eta_t\tau \log Z\\
&=& -\log Z,
\end{eqnarray*}
where we used Jensen's inequality applied with the concave logarithmic function. We deduce
$$\KL(\pi , \pi_t^\mu) \leq \eta_t\tau \KL(\pi , \mu) + (1-\eta_t\tau)\KL(\pi , \pi_{t}) - \eta_t\tau \KL(\pi_t^\mu, \mu).$$
\end{proof}

Now we use Lemma 7 of \cite{MAIO}, restated below with notation.
\begin{lemma}
Let $p\geq 1$ and $q\geq 1$ such that $1/p+1/q=1$. Let $\phi$ be a strongly convex function with respect to the $\ell_p$-norm $\|\cdot \|_p$ with some modulus $\sigma$, i.e., for any $\pi,\pi'$,
\beq \notag
\phi(\pi)\geq \phi(\pi') + \nabla \phi(\pi')\cdot(\pi-\pi')+\frac{\sigma}{2} \|\pi-\pi'\|^2.
\eeq
Write $D_\phi$ the associated Bregman divergence: for $\pi,\pi'$,
$$D_\phi(\pi,\pi')\eqdef \phi(\pi)-\phi(\pi') - \nabla \phi(\pi')\cdot( \pi-\pi').$$
Let $\delta$ be a vector of dimension $|\actionspace|$. For any $\pi^-\in \Delta(\actionspace)$, define $\pi^+$ as
\beq
\pi^+ = \arg\max_{\pi\in\Delta(\actionspace)} \left\lbrack  \sum_y \pi(y) \delta(y) - D_\phi(\pi,\pi^-) \right\rbrack,
\eeq
Then for any $\pi\in \Delta(\actionspace)$, we have, 
\beqan 
D_\phi(\pi,\pi^+) \leq D_\phi(\pi, \pi^-) +  \sum_y (\pi^-(y)-\pi(y)) \delta(y) + (2/\sigma) \|\delta\|_q^2.
\eeqan
\end{lemma}

We apply this lemma with $\pi^+=\pi_{t+1}$ and $\pi^-=\pi_t^\mu$, with the vector $\delta(y) =\eta_t \P(y\succ \pi_t^\mu)$, and as Bregman divergence $D_\phi$ we choose the $\KL$ (which corresponds to the choice of the entropy regularizer $\phi(\pi) = \sum_y \pi(y)\log\pi(y)$). For $p=1$, $q=\infty$, the regularizer $\phi$ is a strongly convex function with respect to the $\ell_1$-norm with a modulus $\sigma=1$; this is a consequence of Pinsker’s inequality, see \cite{InfoTheoryBook}.

We deduce that for any $\pi$,
\begin{eqnarray}\label{eqn:bound.KL} 
\KL(\pi,\pi_{t+1}) \leq \KL(\pi, \pi_{t}^\mu) + \eta_t \sum_y (\pi_{t}^\mu(y)-\pi(y)) \P(y\succ \pi_t^\mu) + 2 \eta_t^2.
\end{eqnarray}

For the choice $\pi=\pi^*_\tau$ and using the previous lemma, we have
\begin{eqnarray*} 
\KL(\pi^*_\tau,\pi_{t+1}) &\leq& \KL(\pi^*_\tau, \pi_{t}^\mu) + \eta_t \sum_y (\pi_{t}^\mu(y)-\pi^*_\tau(y)) \P(y\succ \pi_t^\mu) + 2 \eta_t^2\\
&\leq& (1-\eta_t\tau) \KL(\pi^*_\tau, \pi_t) + \eta_t\tau\left( \KL(\pi^*_\tau, \mu) - \KL(\pi_t^\mu,\mu)\right)\\
& &+
\eta_t \left( \P(\pi_{t}^\mu\succ \pi_t^\mu) - \P(\pi^*_\tau\succ \pi_t^\mu)\right) + 2 \eta_t^2\\
&=& (1-\eta_t\tau) \KL(\pi^*_\tau, \pi_t) + \eta_t\left[ 1/2 - \P(\pi^*_\tau\succ \pi_t^\mu) + \tau \KL(\pi^*_\tau, \mu) - \tau \KL(\pi_t^\mu,\mu)\right] + 2 \eta_t^2\\
&=& (1-\eta_t\tau) \KL(\pi^*_\tau, \pi_t) + \eta_t\left[ 1/2 - \P_\tau(\pi^*_\tau\succ \pi_t^\mu) \right] + 2 \eta_t^2\\
&\leq& (1-\eta_t\tau) \KL(\pi^*_\tau, \pi_t) + 2 \eta_t^2,
\end{eqnarray*}
where the last inequality comes from the fact that $\pi_\tau^*$ is the Nash of the regularized game $\P_\tau$:
$\P_\tau(\pi^*_\tau\succ \pi_t^\mu)\geq \P_\tau(\pi^*_\tau\succ \pi^*_\tau)=1/2$
and the last equality comes from the definition of the regularized preference.

This inequality with $\eta_t=2/(\tau(t+2))$ applied to $t=0$ gives
$$\KL(\pi^*_\tau,\pi_1)\leq \frac{2}{\tau^2}.$$
Then, by induction, assuming $\KL(\pi^*_\tau,\pi_t)\leq \frac{8}{\tau^2(t+1)}$,
\beqan
\KL(\pi^*_\tau,\pi_{t+1}) &\leq&  \left(1-\frac{2}{t+2}\right)\frac{8}{\tau^2(t+1)}+\frac{8}{\tau^2(t+2)^2}\\
&\leq&  \left(1-\frac{2}{t+2}+\frac{1}{t+2}\right)\frac{8}{\tau^2(t+1)}\\
&=& \frac{8}{\tau^2(t+2)}.
\eeqan

\section{Proof of Proposition \ref{prop:unique}}
\label{proof:unique}

The mappings $\pi\mapsto {\cal P}(\pi\succ\pi')$ and $\pi'\mapsto {\cal P}(\pi\succ\pi')$ are linear in $\pi$ (respectively in $\pi'$) thus $\pi\mapsto {\cal P}_\tau(\pi\succ\pi')$ is  concave and $\pi'\mapsto {\cal P}_\tau(\pi\succ\pi')$ is  convex. Existence of a Nash equilibrium is derived from the minimax theorem for convex-concave functions \citep{sion1958general}. 

The uniqueness of the Nash equilibrium essentially relies on the strict convexity/concavity of these mappings. We now give a proof of existence and uniqueness using variational inequalities. 

We first note that since $\P_{\tau}(\pi'\succ \pi)=1-\P_{\tau}(\pi\succ \pi')$ we can re-express the minimax game of Eq.~\ref{eq:minmax.reg} as an antisymmetric two-player game with payoffs of policy $\pi$ and $\pi'$ are defined as 
$$R(\pi;\pi')= \P(\pi\succ\pi')-\tau\KL_{\rho}(\pi,\mu)$$ 
and 
$$R(\pi';\pi)= \P(\pi'\succ\pi)-\tau\KL_{\rho}(\pi',\mu)$$
respectively. First we notice that since the payoff of this  game is  concave in $\pi$ and $\pi'$, it possesses a Nash equilibrium \citep[Theorem 1]{rosen1965existence}.

To show that this game has unique Nash equilibrium  we need to show that its corresponding variational inequality  is  strictly monotone \citep[Theorem 2]{rosen1965existence}. Let $\bar\pi=[\pi,\pi']$  and $v(\bar\pi)=[\nabla_{\pi} R(\pi;\pi'),\nabla_{\pi'} R(\pi';\pi)]$. Then every  Nash equilibrium  of the game should satisfy the following variational inequality for all $\bar \pi$:

$$v^T(\bar\pi^*)(\bar\pi^* -\bar \pi) \leq 0$$

Furthermore the variational inequality is strictly monotone if  and only if for every $\bar\pi_1$ and $\bar \pi_2$ we have that 

\begin{equation}
\label{eq:strict.monotone.def}
(v(\bar\pi_1)-v(\bar \pi_2))^T(\bar\pi_1 -\bar \pi_2) \leq 0
\end{equation}
with equality only holds at $\bar\pi_1=\bar\pi_2$ \citep[Theorem 2]{rosen1965existence}. We can show this inequality holds by expanding the terms on LHS. For every context $x$ let denote $v(\bar\pi)(x)$ as the partial derivative $v(\bar\pi)$ for $x$. We have:

\begin{equation*}
\label{eq:monotone}
v(\bar\pi)(x)=\rho(x)[\mathbf P (y\succ\pi'|x)-\tau\log(\pi/\mu|x)-1,\mathbf P (y\succ\pi|x)-\tau\log(\pi'/\mu|x)-1],   
\end{equation*}
where $\mathbf P (y\succ\pi'|x)=[p(y_i\succ\pi|x)]_{i=1:N}$ and $\log(\pi/\mu|x)=[\log(\pi(y_i|x)/\mu(y_i|x))]_{i=1:N}$, in which $N$ is the size of the  generation set.  Plugging this in the LHS of Eq.\ref{eq:strict.monotone.def} and then exploiting the non-negativity of KL-divergence implies:

\begin{align*}
(v(\bar\pi_1)-v(\bar \pi_2))^T(\bar\pi_1 -\bar \pi_2)&=\underbrace{\P(\pi_1\succ\pi'_1)+\P(\pi'_1\succ\pi_1)+\P(\pi_2\succ\pi'_2)+\P(\pi'_2\succ\pi_2)}_{=2}
\\&
-\underbrace{(\P(\pi_2\succ\pi'_1)+\P(\pi'_1\succ\pi_2)+\P(\pi_1\succ\pi'_2)+\P(\pi'_2\succ\pi_1))}_{=2}
\\&-\tau (\KL_{\rho}(\pi_1||\pi_2)+\KL_{\rho}(\pi_2||\pi_1)+\KL_{\rho}(\pi'_1||\pi'_2)+\KL_{\rho}(\pi'_2||\pi'_1))
\\&
=-\tau (\KL_{\rho}(\pi_1||\pi_2)+\KL_{\rho}(\pi_2||\pi_1)+\KL_{\rho}(\pi'_1||\pi'_2)+\KL_{\rho}(\pi'_2||\pi'_1))\leq 0
\end{align*}
with equality only at $\bar\pi_1=\bar\pi_2$.

\section{Deep Learning Implementation of NLHF}\label{sec:implementations}

Now, building upon the insights from Nash-MD, we explore potential gradient-based algorithms for deep-learning architectures designed for the computation of the Nash equilibrium of a preference model, with a specific focus on their applicability in the context of LLMs.

\subsection{Generating one token at the time, instead of a full sequence}
\label{sec:one.token_at_a_time}

In LLMs it is usually the case that tokens are generated one at a time in an autoregressive manner. Thus the response $y\sim\pi(\cdot|x)$ can be written as $y=y_{0:N}$ (where $y_{0:N}\eqdef (y_0,\dots, y_N)$), where each token $y_{n}$ is generated from a distribution $\pi(\cdot|x, y_{0:n-1})$ conditioned on previous tokens, such that $\pi(y_{0:N}|x) = \prod_{n=0}^{N} \pi(y_n|x, y_{0:n-1})$. 
In practice (see the experiments section for results on LLMs) we will implement this token-per-token autoregressive generation of responses $y\sim\pi(\cdot|x)$ using next token distributions (implemented as a softmax over logits).

Now consider a parametric policy $\pi_\theta$. Nash-MD requires the generation of alternative responses $y'\sim \pi_{\theta}^\beta(y|x)$ sampled from the regularized policy $\pi_{\theta}^\beta(y|x)\propto (\pi_{\theta}(y|x))^{1-\beta}(\mu(y|x))^\beta$ which is defined like in \eqref{eq:geometric-mixture} as a geometric mixture between the current policy $\pi_{\theta}$ and the reference policy $\mu$. However it is not easy to generate a sequence $y$ from this distribution by sampling one token $y_n$ at a time. In particular, since $\pi_{\theta}^\beta$ is not a simple (arithmetic) mixture, we cannot select one policy $\pi_{\theta}$ or $\mu$ according to some prior probability (that would depend on the mixing parameter $\beta$) and then generate a sequence of tokens following that policy. Additionally, defining the normalization constant $c$ as in \eqref{eq:Nash.MD.log} for the full mixture $\pi_{\theta}^\beta$ is computationally prohibitive given the large number of possible sequences; instead, we would like to proceed by generating a token at a time. The approach we follow in our experiments consists in generating a token $y_n$ from the marginal (geometric) mixture $\tilde \pi_{\theta}^\beta(\cdot|x, y_{0:n-1})$ defined by
$$\log \tilde \pi_{\theta}^\beta(y_n|x, y_{0:n-1}) = (1-\beta) \log \pi_{\theta}(y_n|x, y_{0:n-1}) + \beta \log \mu(y_n|x, y_{0:n-1})+C(x, y_{0:n-1}),$$
where the normalization constant $C$ depends on $x, y_{0:n-1}$. In order to sample from this marginal geometric mixture over the $n$th token, we evaluate the corresponding logits of both the current policy $\pi_{\theta}$ and the reference policy $\mu$ (conditioned on $(x, y_{0:n-1})$), we compute their ($\beta$-arithmetic) mixture, and sample a next token $y_{n}$ from the corresponding softmax distribution. 
We call this corresponding product of marginal (geometric) mixtures over individual tokens the {\bf one-step-at-a-time regularized policy}
\begin{align*}
    \tilde \pi^\beta_{\theta}(y|x) \eqdef \prod_{n=0}^N \tilde \pi_{\theta}^\beta(y_n|x, y_{0:n-1}).
\end{align*}
Notice that the one-step-at-a-time regularized policy $\tilde \pi^\beta_{\theta}(y|x)$ is different from the original regularized policy $\pi_{\theta}^\beta(y|x)$ because the sequence of normalization constants $C(x, y_{0:n-1})$ depend on the specific sample path $y_{0:n-1}$ and does not necessarily correspond to the full normalization constant $c$ defined in \eqref{eq:Nash.MD.log}. We leave the  analysis of the difference between these two policies for future work.

\subsection{Computing the Nash equilibrium using regularized policy gradient}\label{sec:PG}

Our  general algorithm for computing the Nash equilibrium of the preference model consists in repeating these steps: 
\begin{itemize}
    \item We randomly select a prompt $x\sim\rho$. 
    \item We generate two responses $y$ and $y'$ (in an autoregressive fashion in the case of LLMs):
    \begin{itemize}
        \item the first one $y\sim\pi_\theta(\cdot|x)$ by following the {\em current policy $\pi_\theta$} that is being optimized;
        \item the second one $y'\sim \pi'(\cdot|x)$ by following an {\em alternative policy $\pi'$}. 
    \end{itemize}
The choice of the alternative policy $\pi'$ that we use for the second generated sample $y'$ depends on the specific algorithm we consider (the description of which is given in the next subsection). 
    \item We update the parameter $\theta$ of the policy $\pi_\theta$ in the direction of the gradient $\nabla_\theta \P_\tau(\pi_\theta\succ \pi')$ of the regularized preference model $\P_\tau$. 
\end{itemize}
We consider two cases, depending on whether a preference model is learnt or not.

\paragraph{$\P$-model-based approach.}
If we have learnt a preference model $\P$ (see Section~\ref{sec:pref.and.reward.models} for example for how one can learn a preference model) we query it to get the preference reward $\P(y\succ y'|x)$ and update $\theta$ by moving it in the direction of the policy gradient estimate
\beq\label{eq:PG}
\hat g(x,y,y') \eqdef \nabla_\theta \log\pi_\theta(y|x) \left( \P(y\succ y'|x)-1/2 -\tau \log\frac{\pi_\theta(y|x)}{\mu(y|x)}\right).
\eeq
Notice we have subtracted the baseline $1/2 = \P(y\succ y|x)$ from the preference $\P(y\succ y'|x)$ (which does not change the expectation of the gradient) as a variance reduction technique that does not require learning a value function as baseline. In practice, when the response $y$ comprises a sequence of tokens $y_{0:N}$, a sample-based estimator to the KL based on the sample response $y$ can be used. Further, this can be decomposed into a sum across token indicies of per-token KL estimators, and the standard policy-gradient variance-reduction trick of only multiplying $\nabla_\theta \log \pi_\theta(y_{n}|x, y_{0:n-1})$ by KL estimator terms corresponding to indices at least as great as $n$ can be applied.

\paragraph{$\P$-model-free approach.}
In the case the preference model $\P(y\succ y'|x)$ comes directly from human preferences: $\P(y\succ y'|x) = \Pr_{Z\sim\nu}(\mbox{Human } Z \mbox{ prefers } y \mbox{ over } y'\mbox{ given } x)$, where $\nu$ is a distribution over humans, and if humans are immediately available to express their preference between any two responses, we can directly estimate the gradient by replacing $\P(y\succ y'|x)$ with $\1{\mbox{Human } Z\mbox{ prefers } y \mbox{ over } y'\mbox{ given } x}$ in \eqref{eq:PG}.
This estimate does not require to learn a preference model first and is thus not affected by possible bias coming from an approximate model. Implementation-wise it requires having access to humans preference immediately after having generated the responses $y$ and $y'$. 

In both model-based and model-free approaches, we have that
\begin{align}\label{eq:PG.algo}
    \nabla_\theta {\cal P}_{\tau}(\pi_\theta\succ\underline{\pi'})
    &= \E_{x\sim\rho,{\left\{ y\sim\pi_\theta(\cdot|x) \atop y'\sim\pi'(\cdot|x)\right. }} \left[\hat g(x,y,y')\right],
\end{align}
(where $\underline{\pi'}$ denotes a stop-gradient on $\pi'$ in the case $\pi'$ would depend on $\theta$).

\subsection{Choice of the alternative policy $\pi'$}\label{sec:alternative.policy}
Now, for the choice of alternative policies $\pi'$ that are used to generate the second sample $y'$, we will consider two different algorithms {\bf Nash-MD-PG} and {\bf Nash-EMA-PG}, that are inspired by, respectively, the mirror-ascent algorithm Nash-MD introduced in the previous section, and a generalization of fictitious play where we consider an exponential moving average. 

\paragraph{Nash-MD-PG.} We define the alternative policy $\pi'=\pi_\theta^{\beta}$ as a geometric-mixture between $\pi_\theta$ and $\mu$ in a similar way as the regularized policy is defined in \eqref{eq:geometric-mixture}: 
\beq\label{eq:nash.md_pg}
\log \pi_\theta^{\beta}(y|x)\eqdef (1-\beta)\log (\pi_\theta(y|x))+\beta\log(\mu(y|x))+c(x),
\eeq
where $\beta\in[0,1]$ is the parameter of the mixture, and $c(x)$ is a constant independent of $y$. This is inspired by the Nash-MD algorithm described in Section~\ref{sec:Nash-MD}, which we have proven to be convergent in Theorem~\ref{thm:MD}. In the case of sequential generation of tokens in LLMs, we apply the one-step-at-a-time version $\tilde \pi_\theta^{\beta}$ of this regularized policy $\pi_\theta^{\beta}$ as defined in Subsection~\ref{sec:one.token_at_a_time}.  
However, the corresponding PG version outlined in Subsection \ref{sec:PG} differs from Nash-MD as defined in Section~\ref{sec:Nash-MD} in a number of ways. 

In addition to using a parametric representation of policies instead of a tabular one, it differs from the fact that it is not directly implementing a mirror descent algorithm but a simple gradient descent on the regularized preference model. In a sense this algorithm is only making a gradient step for the inner optimization problem of \eqref{eq:Nash.MD}, whereas a more faithful variant of Nash-MD would use a two-time scale algorithm and perform several gradient steps (while keeping $\pi_\theta$ and $\pi_\theta^{\beta}$ fixed) until the inner loop has reached an optimum, before updating $\pi_\theta$ and $\pi_\theta^{\beta}$. Another apparent difference is that Nash-MD uses a KL-regularization w.r.t.~the mixture policy $\pi_\theta^{\beta}$, whereas Nash-MD-PG uses a KL w.r.t.~the reference policy $\mu$. However, we have that $$\KL(\pi_\theta,\pi_\theta^{\beta}) = (1-\beta) \KL(\pi_\theta,\pi_\theta) + \beta \KL(\pi_\theta,\mu)-\E_{x\sim\rho}[c(x)]=\beta \KL(\pi_\theta,\mu)-\E_{x\sim\rho}[c(x)],$$ where $c(x)$ is the normalizing constant in \eqref{eq:nash.md_pg}. Thus, we have
$$\nabla_\theta \KL(\pi_\theta,\underline{\pi_\theta^{\beta}}) = \beta \nabla_\theta \KL(\pi_\theta,\mu) \, ,$$ and since we perform a single step of gradient descent before updating $\pi_\theta$, regularizing with respect to the mixture $\pi_\theta^{\beta}$ (in Nash-MD) is equivalent to regularizing w.r.t.~$\mu$ (in Nash-MD-PG). Further, we use an additional parameter $\beta$ (to define the mixture) that can be further tuned independently of $\tau$. 

Thus, while it is possible to implement Nash-MD more faithfully, such as by incorporating two-timescale policy gradient versions or exploring variants of regularized policy gradient methods such as PPO \citep{PPO} or NeuRD \citep{NeuRD}, we contend that the essence of Nash-MD is encapsulated in Nash-MD-PG for the following reason: the policy gradient algorithm \eqref{eq:PG.algo} improves the current policy $\pi_\theta$ by playing against the geometric mixture $\pi_\theta^{\beta}$ while preserving regularization with respect to $\pi_\theta^{\beta}$.

\paragraph{Extreme cases for $\beta\in[0,1]$.}
Consider the alternative policy $\pi_\theta^\beta$ of Nash-MD-PG when $\beta\in[0,1]$ takes its extreme possible values: $\beta=0$ or $1$. When $\beta=0$ then $\pi_\theta^{\beta=0}=\pi_\theta$, thus the alternative policy is the current policy, and this algorithm is simply a version of {\bf self-play} (SP) where one improves its policy by playing against oneself. We do not expect this algorithm (even in its tabular form) to enjoy a last-iterate convergence to the Nash equilibrium; see the discussion around the OMD algorithm in \eqref{eq:OMD.short}.

Now, when $\beta=1$, then the alternative policy is $\pi_\theta^{\beta=1}=\mu$, thus we are improving the current policy against the (fixed) reference policy $\mu$ (i.e., optimizing $\pi\mapsto\P_\tau(\pi,\mu)$), thus this a version of {\bf best-response} (BR) against $\mu$. This will generally not converge to the Nash equilibrium either because there is no reason that this BR cannot be exploited.

\paragraph{Nash-EMA-PG.} As an alternative to Nash-MD-PG, we consider as alternative policy $\pi'$ another mixture policy $\pi'\eqdef \pi_{\bar\theta_t}$ where $\bar\theta_t$ is a exponential moving average (EMA) of the past values of the parameter $(\theta_s)_{s\leq t}$, defined (recursively) by $\bar \theta_t = (1-\beta) \theta_{t} + \beta \theta_0$. Thus when $\beta=0$ then $\pi_{\bar\theta_t}=\pi_{\theta_t}$ and the algorithm is just self-play, and when $\beta=1$, then $\pi_{\bar\theta_t}=\pi_{\theta_0}$ and the algorithm is a best response again the fixed initial policy $\pi_{\theta_0}$.

Now for any other $\beta\in(0,1)$ the policy uses as parameter a mixture of past parameters. Because of the non-linearity of the policy representation, there is no guarantee that this policy is the mixture of the corresponding past policies. However, prior work in deep learning \citep{BYOL,Wortsman2022ModelSA,busbridge2023scale,rame2023rewarded}  suggests that it could be a reasonable first-order approximation to it.

\section{Experiments}\label{apx:experiments}

We now report experiments on a summarisation task and compare several algorithms for NLHF (self-play, best-response against $\mu$, Nash-MD-PG and Nash-EMA-PG) as well as a RLHF baseline.

\subsection{Preference models versus reward models}\label{sec:pref.and.reward.models}

In this section, we compare parametric preference models $\P_\theta$ and reward models $r_\theta$. Preference models assigns a score $\P_\theta(y\succ y'|x)\in [0,1]$ that can be interpreted as the probability of generation $y$ being preferred to generation $y'$ given the context $x$.
The preference  $\P_\theta(y\succ y'|x)$ is initialised by using a LLM prompted in the following way:

\texttt{You are an expert summary rater. Given a piece of text and two of its possible summaries, output 1 or 2 to indicate which summary is better.
Text - $\langle \mbox{text}\rangle$, Summary 1 - $\langle \mbox{summary1}\rangle$, Summary 2 - $\langle \mbox{summary2}\rangle$.
Preferred Summary -}

where \texttt{$\langle \mbox{text}\rangle$} corresponds to $x$, \texttt{$\langle \mbox{summary1}\rangle$} to $y$, and \texttt{$\langle \mbox{summary2}\rangle$} to $y'$. We then use the last logit for an arbitrary chosen token and pass it through a sigmoid function to output a single number in $[0,1]$. This number models the preference  $\P_\theta(y\succ y'|x)$. 
We train the LLM to fit the underlying human preference probability $\P(y\succ y'|x)$ by minimizing a cross-entropy loss on a dataset $D=\{(x^k, y^k_w, y^k_l)_{1\leq k\leq K}\}$, where $y^k_w$ is the preferred generation, $y^k_l$ is the less preferred generation and $K$ is the number of examples:
\begin{equation*}
    \mathcal{L}_\P(\theta) = -\mathbb{E}_{(x, y_w, y_l)\sim D}\left[ \log\left(\P_\theta(y_w \succ y_l|x)\right)\right].
\end{equation*}
Reward models assigns a score $r_\theta(x,y)\in\mathbb{R}$ that can be interpreted as the value of a generation $y$ given a context $x$. 
The reward  $r_\theta(y|x)$ is defined by prompting the LLM in the following way:
\texttt{`Context - $\langle \mbox{text}\rangle$, Summary - $\langle \mbox{summary}\rangle$'} where \texttt{$\langle \mbox{text}\rangle$} corresponds to $x$ and \texttt{$\langle \mbox{summary}\rangle$} to $y$. We then use the last logit for an arbitrary chosen token to output a single number. This number models the reward $r_\theta(y|x)$. Reward models are trained to fit the underlying human preference probability $\P(y\succ y'|x)$ via a Bradley-Terry model $\P_{BT}(y\succ y'|x) \eqdef \sigma\left(r_\theta(x,y)-r_\theta(x,y')\right)$ where $\sigma(x)$ is the sigmoid function. They use the same preference dataset $D$ and minimize the following cross-entropy loss:
\begin{equation*}
    \mathcal{L}_r(\theta) = -\mathbb{E}_{(x, y_w, y_l)\sim D}\left[ \log\left(\sigma\left(r_\theta(y_w|x)-r_\theta(y_l|x)\right)\right)\right].
\end{equation*}
In our experiments, we use the summarization dataset described in~\cite{stiennon2020learning} that has been built from the TL;DR dataset~\citep{TLDR}.
We train our preference and reward models on the train set $D_{\texttt{Train}}$, that contains $92820$ examples,  and evaluate them on a test set of high confidence data $D_{\texttt{Test}}$. To measure the quality of our models we use the expected agreement, also called {\em accuracy}, between our models and the human ratings:
\begin{align*}
\mathcal{A}(\P_\theta) &= \mathbb{E}_{(x, y_w, y_l)\sim D}\left[ \mathbf{1}_{\{\P_\theta(y_w \succ y_l|x)\geq0.5\}}\right],
\\
\mathcal{A}(r_\theta) &= \mathbb{E}_{(x, y_w, y_l)\sim D}\left[ \mathbf{1}_{\{\sigma\left(r_\theta(y_w|x)-r_\theta(y_l|x)\right)\geq0.5\}}\right].
\end{align*}

\begin{figure}[ht]
    \centerline{
    \hfill
    \includegraphics[width=0.45\textwidth]{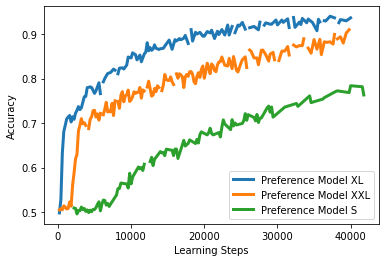}
    \hfill
    \includegraphics[width=0.45\textwidth]{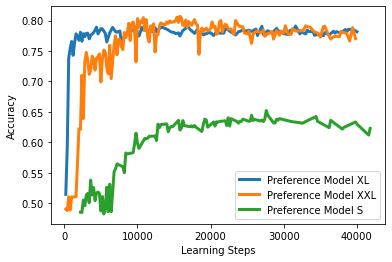}
    \hfill
    }
    \caption{Learning curves showing the accuracy of preference models of different sizes on the train set (left) and on the test set (right).}
    \label{fig:Preference Model Training}
\end{figure}

Our first experiment (see Figure~\ref{fig:Preference Model Training}) shows the accuracy of preference models with different sizes. Our models are T5X encoder-decoder models (transformer models) that have been described in detail in \citep{roberts2203scaling, roit2023factually}. We use different sizes: T5X-small (110M), T5X-XL (3B) and T5X-XXL (11B). We see, on the test set, that the bigger the model the better the accuracy. However, there is relatively small gains going from 3B to 11B in this specific summarization task. In the remaining, we therefore run our experiments on T5X-XL models only.

Our second experiment consists in looking at the accuracy of T5X-XL reward model versus the accuracy of a T5X-XL preference model. We observe that the preference model has a slightly better accuracy than the reward model on the test set (peak accuracy for the preference model is around $0.78$ vs $0.76$ for the reward model). 

\begin{figure}[htbp]
    \centerline{
    \hfill
    \includegraphics[width=0.45\textwidth]{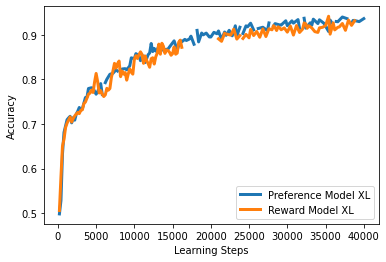}
    \hfill
    \includegraphics[width=0.45\textwidth]{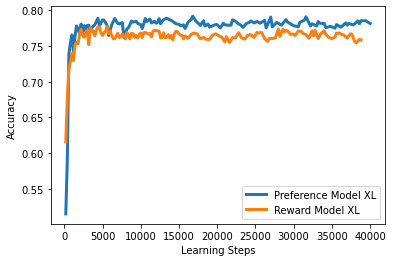}
    \hfill
    }
    \caption{Learning curves showing the accuracy of a preference model versus the accuracy of a reward model of the same size on the train set (left) and on the test set (right).}
    \label{fig:reward vs preference Model Training}
\end{figure}

\subsection{Supervised fine-tuned (SFT) initial policy}
\label{sec:SFT}

In all our experiments, we will initialize our policy with a T5X-L model and fine-tune it by supervised learning using the OpenAI dataset described in \cite{stiennon2020learning} that was built from the TL;DR dataset~\citep{TLDR}. We call this supervised fine-tuned model the SFT. In all our experiments, our policies are initialized with this SFT.

For all our policy models, we opted for a T5X-L model, as opposed to T5X-XL, for computational efficiency and to compute the pairwise comparisons across our policies. The primary objective of these experiments is to provide a proof of concept for the NLHF approach introduced in this paper, rather than striving for state-of-the-art performance in text summarization. Therefore, our aim is to conduct a fair and equitable comparison among the various approaches.

\subsection{RLHF baseline}\label{sec:RLHF}
We established a RLHF baseline by initializing our model with the SFT and then updating the policy by doing 10000 steps of a regularized policy gradient update:
\beq
\label{eq:reg-pg}
\E_{x\sim\rho, y\sim\pi_\theta(\cdot|x)}\left[\nabla_\theta \log\pi_\theta(y|x) \left( R(x,y) -\tau \KL(\pi_\theta(\cdot|x), \piref(\cdot|x))\right)\right],
\eeq
where the reward $R(x,y)$ comes from the trained T5X-XL reward model, as described in Subsection~\ref{sec:pref.and.reward.models}. We conducted a sweep across a set of values ${0.01, 0.02, 0.05, 0.1, 0.2}$ for the parameter $\tau$ of the KL-regularization. The value $\tau=0.05$ has been selected for the pairwise comparison table below.

\subsection{NLHF algorithms Nash-MD and Nash-EMA}
\label{sec:NLHF}
We initialize our policy with the SFT and update the model by executing the Nash-MD-PG and Nash-EMA-PG algorithms as outlined in Section~\ref{sec:implementations}. The preference model $\P$ used in these algorithms is derived from the trained T5X-XL model, as described in Subsection~\ref{sec:pref.and.reward.models}.

We conducted a sweep over the values $\tau\in \{0.02, 0.01, 0.008, 0.005\}$ and selected $\tau=0.008$ for all Nash-MD and Nash-EMA experiments for the pairwise comparison table below.

For Nash-MD-PG we conducted a sweep over the mixing coefficient $\beta \in \{0, 0.125, 0.250, 0.375, 0.5,$ $0.625, 0.75, 0.875, 1.0\}$ (used in the definition of the alternative policy defined in Section~\ref{sec:alternative.policy}) and for Nash-EMA-PG we have swept over $\beta\in\{0, 0.999, 0.9995, 0.9999, 1.0\}$.

\subsection{Pairwise preference between all the models}
Here are the list of all the models we considered for pairwise preference comparison.
\begin{itemize}
    \item SFT: Supervised-fined-tuned, described in Subsection~\ref{sec:SFT}. All models all initialised with this SFT and this SFT is also the policy $\mu$ we use for the \KL-regularization.
    \item RLHF described in Subsection~\ref{sec:RLHF} with regularization coefficient $\tau=0.05$.
    \item SP (self-play). This corresponds to Nash-MD-PG with mixture coefficient $\beta=0$ (or equivalently Nash-EMA-PG with $\beta=0$ as both algorithms are equivalent for $\beta=0$), described in Subsection~\ref{sec:NLHF}. The policy improves by playing against itself (the alternative policy $\pi'=\pi_\theta$ is the current policy).
    \item MD1 to MD6 is Nash-MD-PG with $\beta\in\{0.125, 0.25, 0.375, 0.5, 0.625, 0.75\}$. 
    \item BR is best-response against SFT. This corresponds to Nash-MD-PG with $\beta=1$ (or equivalently Nash-EMA-PG with $\beta=1$). The policy improves by playing against the fixed SFT policy.
    \item EMA1 and EMA2 are the last-iterate of Nash-EMA-PG (i.e., returns the last policy), with $\beta\in\{0.999, 0.9995\}$. 
    \item EMA1* and EMA* are the EMA policy of Nash-EMA-PG (i.e., returns the policy with average weight) with $\beta\in\{0.999,0.9995\}$. 
\end{itemize}

All models are trained for $10000$ steps. The Nash-MD models (as well as SP and BR) and Nash-EMA are trained with a regularization coefficient of $\tau=0.008$. The pairwise preference comparisons under $\P_\tau$ are given in Table~\ref{results:pref}; these figures are estimated based on 1,000 pairwise comparisons, and hence an upper bound on the width of a  95\% confidence interval for each is $\pm 0.032$, based on the exact Clopper-Pearson method for Bernoulli proportions \citep{clopper1934use}. Note that the Clopper-Pearson method can be used to deduce a per-element confidence interval which may be considerably narrower in cases where the empirically observed preference rate is close to 0 or 1.

\begin{table}[ht]
\caption{The regularized preference $\P_\tau(\pi_c\succ\pi_r)$   between column policy $\pi_c$ against row policy $\pi_r$}
\label{results:pref}
\centering
{\scriptsize
\setlength\tabcolsep{4.7pt}
\begin{tabular}{c|l|l|l|l|l|l|l|l|l|l|l|l|l|l}
$\P_\tau$ &   SFT &  RLHF &    SP &   {\bf MD1} &   MD2 &   MD3 &   MD4 &   MD5 &   MD6 &    BR &  EMA1 &  EMA2 &  EMA1* &  EMA2* \\
\midrule
      SFT & 0.500 & 0.975 & 0.981 & {\bf 0.986} & 0.983 & 0.982 & 0.979 & 0.970 & 0.967 & 0.933 & 0.965 & 0.970 &  0.971 &  0.975 \\
     RLHF & 0.025 & 0.500 & 0.741 & {\bf 0.769} & 0.752 & 0.744 & 0.661 & 0.450 & 0.340 & 0.167 & 0.640 & 0.531 &  0.617 &  0.671 \\
       SP & 0.019 & 0.259 & 0.500 & {\bf 0.547} & 0.506 & 0.509 & 0.406 & 0.244 & 0.185 & 0.082 & 0.418 & 0.338 &  0.363 &  0.450 \\
      MD1 & 0.014 & 0.231 & 0.453 & {\bf 0.500} & 0.471 & 0.469 & 0.354 & 0.224 & 0.165 & 0.079 & 0.372 & 0.308 &  0.348 &  0.409 \\
      MD2 & 0.017 & 0.248 & 0.494 & {\bf 0.529} & 0.500 & 0.492 & 0.393 & 0.231 & 0.182 & 0.084 & 0.426 & 0.315 &  0.375 &  0.454 \\
      MD3 & 0.018 & 0.256 & 0.491 & {\bf 0.531} & 0.508 & 0.500 & 0.380 & 0.230 & 0.153 & 0.087 & 0.411 & 0.328 &  0.349 &  0.457 \\
      MD4 & 0.021 & 0.339 & 0.594 & {\bf 0.646} & 0.607 & 0.620 & 0.500 & 0.306 & 0.224 & 0.088 & 0.508 & 0.416 &  0.458 &  0.531 \\
      MD5 & 0.030 & 0.550 & 0.756 & {\bf 0.776} & 0.769 & 0.770 & 0.694 & 0.500 & 0.380 & 0.169 & 0.682 & 0.554 &  0.627 &  0.697 \\
      MD6 & 0.033 & 0.660 & 0.815 & {\bf 0.835} & 0.818 & 0.847 & 0.776 & 0.620 & 0.500 & 0.269 & 0.735 & 0.644 &  0.706 &  0.777 \\
       BR & 0.067 & 0.833 & 0.918 & {\bf 0.921} & 0.916 & 0.913 & 0.912 & 0.831 & 0.731 & 0.500 & 0.856 & 0.789 &  0.830 &  0.875 \\
     EMA1 & 0.035 & 0.360 & 0.582 & {\bf 0.628} & 0.574 & 0.589 & 0.492 & 0.318 & 0.265 & 0.144 & 0.500 & 0.407 &  0.448 &  0.507 \\
     EMA2 & 0.030 & 0.469 & 0.662 & {\bf 0.692} & 0.685 & 0.672 & 0.584 & 0.446 & 0.356 & 0.211 & 0.593 & 0.500 &  0.540 &  0.627 \\
    EMA1* & 0.029 & 0.383 & 0.637 & {\bf 0.652} & 0.625 & 0.651 & 0.542 & 0.373 & 0.294 & 0.170 & 0.552 & 0.460 &  0.500 &  0.589 \\
    EMA2* & 0.025 & 0.329 & 0.550 & {\bf 0.591} & 0.546 & 0.543 & 0.469 & 0.303 & 0.223 & 0.125 & 0.493 & 0.373 &  0.411 &  0.500 \\
\end{tabular}
}
\end{table}

We will analyse these results after the next section where we describe an evaluation of our models based on a preference model build from a much larger LLM.

\subsection{Evaluation using the PaLM 2 preference model}
While the ideal approach for evaluating our models would involve soliciting human preferences between summaries generated by different models, we resort to a proxy method using the highly capable LLM, PaLM 2 Large \citep{anil2023palm}. We query this model to obtain a preference signal, which we refer to as the PaLM 2 preference model $\P^*(y\succ y'|x)$, achieved by prompting the LLM in the following manner:
\begin{align*}
    &\texttt{`You are an expert summary rater. Given a piece of text and two of its}\\
    &\texttt{possible summaries, output 1 or 2 to indicate which summary is better. }\\
    &\texttt{Text - $\langle \mbox{text}\rangle$, Summary 1 - $\langle \mbox{summary1}\rangle$, Summary 2 - $\langle \mbox{summary2}\rangle$.}\\
    &\texttt{Preferred Summary -',}    
\end{align*}
where \texttt{$\langle \mbox{text}\rangle$} corresponds to $x$, \texttt{$\langle \mbox{summary1}\rangle$} to $y$, and \texttt{$\langle \mbox{summary2}\rangle$} to $y'$. 

This evaluation approach shares similarities with the method employed by \citet{lee2023rlaif}. To obtain an assessment of the preference $\P^*(\pi\succ \pi')$, we compute the ratio between the total number of token '1' generated and the total number of token '1' or '2' across $2000$ samples drawn from the distribution $(x\sim\rho,y\sim\pi(\cdot|x),y'\sim\pi'(\cdot|x))$.

This $\P^*$ serves as an approximate surrogate for human preferences. Notably, it is essential to highlight that the preference model $\P$ utilized during the training of our policies is considerably smaller in size than $\P^*$ and corresponds to a different model. Specifically, $\P$ is based on the T5X-XL model, fine-tuned with TL;DR data, whereas $\P^*$ is derived from the PaLM 2 Large model.

The pairwise preference comparisons under $\P^*$ using the PaLM 2 Large model are given in Table~\ref{results:p*}. As each element is estimated with $2000$ samples, the confidence interval, an upper bound on the 95\% confidence interval is given by $\pm 0.023$, based on the exact Clopper-Pearson method for Bernoulli proportions \citep{clopper1934use}.

\begin{table}[ht]
\caption{PaLM 2 preference $\P^*(\pi_c\succ\pi_r)$ model between column policy $\pi_c$ against row policy $\pi_r$.}
\label{results:p*}
\centering
{\scriptsize
\setlength\tabcolsep{4.7pt}
\begin{tabular}{c|l|l|l|l|l|l|l|l|l|l|l|l|l|l}
$\P^*$ &   SFT &  RLHF &    SP &   {\bf MD1} &   MD2 &   MD3 &   MD4 &   MD5 &   MD6 &    BR &  EMA1 &  EMA2 &  EMA1* &  EMA2* \\
\midrule
  SFT & 0.500 & 0.990 & 0.983 & {\bf 0.982} & 0.989 & 0.987 & 0.985 & 0.982 & 0.965 & 0.943 & 0.970 & 0.961 &  0.977 &  0.980 \\ \cline{4-11}
  RLHF & 0.010 & 0.500 & \multicolumn{1}{l|}{{\color{blue}{\bf 0.489}}} &
  \multicolumn{1}{l|}{{\color{blue}{\bf 0.598}}} &
  \multicolumn{1}{l|}{{\color{blue}{\bf 0.519}}} &
  \multicolumn{1}{l|}{{\color{blue}{\bf 0.561}}} & \multicolumn{1}{l|}{{\color{blue}{\bf 0.501}}} & \multicolumn{1}{l|}{{\color{blue}{\bf 0.436}}} & \multicolumn{1}{l|}{{\color{blue}{\bf 0.284}}} &
  \multicolumn{1}{l|}{{\color{blue}{\bf 0.148}}} & 
  0.468 & 0.320 &  0.477 &  0.510 \\ \cline{4-11}
    SP & 0.017 & 0.511 & 0.500 & {\bf 0.592} & 0.504 & 0.545 & 0.499 & 0.451 & 0.310 & 0.211 & 0.445 & 0.362 &  0.464 &  0.488 \\
   MD1 & 0.018 & 0.402 & 0.408 & {\bf 0.500} & 0.425 & 0.470 & 0.369 & 0.362 & 0.238 & 0.163 & 0.391 & 0.270 &  0.400 &  0.447 \\
   MD2 & 0.011 & 0.481 & 0.496 & {\bf 0.575} & 0.500 & 0.513 & 0.491 & 0.434 & 0.298 & 0.196 & 0.460 & 0.351 &  0.430 &  0.496 \\
   MD3 & 0.013 & 0.439 & 0.455 & {\bf 0.530} & 0.487 & 0.500 & 0.484 & 0.408 & 0.273 & 0.187 & 0.429 & 0.323 &  0.413 &  0.472 \\
   MD4 & 0.015 & 0.499 & 0.501 & {\bf 0.631} & 0.509 & 0.516 & 0.500 & 0.428 & 0.265 & 0.161 & 0.468 & 0.358 &  0.437 &  0.503 \\
   MD5 & 0.018 & 0.564 & 0.549 & {\bf 0.638} & 0.566 & 0.592 & 0.572 & 0.500 & 0.329 & 0.210 & 0.532 & 0.389 &  0.518 &  0.539 \\
   MD6 & 0.035 & 0.716 & 0.690 & {\bf 0.762} & 0.702 & 0.727 & 0.735 & 0.671 & 0.500 & 0.342 & 0.652 & 0.548 &  0.651 &  0.691 \\
    BR & 0.057 & 0.852 & 0.789 & {\bf 0.837} & 0.804 & 0.813 & 0.839 & 0.790 & 0.658 & 0.500 & 0.743 & 0.640 &  0.752 &  0.774 \\
  EMA1 & 0.030 & 0.532 & 0.555 & {\bf 0.609} & 0.540 & 0.571 & 0.532 & 0.468 & 0.348 & 0.257 & 0.500 & 0.381 &  0.480 &  0.556 \\
  EMA2 & 0.039 & 0.680 & 0.638 & {\bf 0.730} & 0.649 & 0.677 & 0.642 & 0.611 & 0.452 & 0.360 & 0.619 & 0.500 &  0.585 &  0.659 \\
 EMA1* & 0.023 & 0.523 & 0.536 & {\bf 0.600} & 0.570 & 0.587 & 0.563 & 0.482 & 0.349 & 0.248 & 0.520 & 0.415 &  0.500 &  0.555 \\
 EMA2* & 0.020 & 0.490 & 0.512 & {\bf 0.553} & 0.504 & 0.528 & 0.497 & 0.461 & 0.309 & 0.226 & 0.444 & 0.341 &  0.445 &  0.500 \\
\end{tabular}
}
\end{table}

\subsection{Analysis of the results}
First, let us mention that the RLHF baseline that we have built is a very strong baseline. It beats SFT with a win rate of $99\%$ marking the highest win rate observed against SFT among all models when using the PaLM 2 preference model $\P^*$

Best-response against self-play (BR) does not exhibit strong performance. Despite being trained explicitly to outperform self-play during training, its $\P^*$-evaluation yields a relatively modest score of $94\%$ against self-play. Furthermore, BR performs poorly against RLHF and all other Nash-based approaches. This suggests the possibility of 'preference hacking,' where BR may be overly adapting to the preference model by overfitting to the specific SFT policy.

Self-play (SP) exhibits strong overall performance, with notable exceptions in the $\P^*$ evaluation against RLHF and the Nash-MD models (for $\beta\leq 0.5$). This suggests that enhancing one's policy through self-play could be a promising avenue for improving the initial model. However, it's essential to acknowledge that self-play does not guarantee the attainment of a Nash equilibrium, as cyclic patterns are possible, as discussed in the Theory Section. In particular, SP is found to be vulnerable to exploitation by certain Nash-MD models.

The Nash-MD models, especially those with $\beta\leq 0.5$, exhibit very strong performance. Notably, Nash-MD models with $\beta=0.125$, $\beta=0.25$, and $\beta=0.375$ outperform all other models, including RLHF. Among them, Nash-MD with $\beta=0.125$ (highlighted in bold as 'MD1') emerges as the top-performing model, surpassing all others in both the training preference model $\P_\tau$ and the evaluation model $\P^*$.

All Nash-EMA models, including EMA1 and EMA2 (representing the last iterate) as well as EMA1* and EMA2* (representing the average policy), are outperformed by Nash-MD for $\beta\leq 0.5$ and RLHF. This observation may suggest that the first-order approximation of the mixture policy as the policy having an average (EMA) weight may not be well-suited in this context, potentially contributing to the overall lower performance.

Examining Nash-MD, which emerges as the most efficient method, it is interesting to note that both extreme values of the mixing parameter $\beta\in[0,1]$, namely $\beta=0$ (self-play) and $\beta=1$ (best-response against SFT), result in suboptimal performance compared to intermediate values of $\beta$ (particularly $\beta=0.125$, $\beta=0.25$, and $\beta=0.375$). This trend is visible, for instance, in the highlighted blue row showing Nash-MD (for $\beta\in{0, 0.125, 0.25, 0.375, 0.5, 0.625, 0.75, 1.0}$) against RLHF. It suggests that improving one's policy by playing against a mixture of the initial policy and the current policy yields superior model improvement compared to interactions with either the initial policy or the current policy in isolation.

\end{document}